\documentclass[10pt,twocolumn,letterpaper]{article}

\usepackage[pagenumbers]{wacv}

\usepackage{array}
\usepackage{adjustbox}
\usepackage{tabularx}
\usepackage{algorithm}
\usepackage[noend]{algpseudocode}
\usepackage{colortbl}
\usepackage{enumitem}
\usepackage{float}
\usepackage{makecell}
\usepackage{placeins}


\newcommand{\method}{FracEvent}
\newcommand{\thetaon}{\theta_{\mathrm{on}}}
\newcommand{\thetaoff}{\theta_{\mathrm{off}}}
\newcommand{\events}{\mathcal{E}}
\definecolor{TableHighlight}{RGB}{232,250,243}
\definecolor{EvalStream}{RGB}{232,250,243}
\definecolor{EvalRecon}{RGB}{248,249,250}
\definecolor{EvalFlow}{RGB}{232,250,243}

\definecolor{TableBlock}{RGB}{235,238,240}

\newcommand{\best}[1]{\textbf{#1}}
\newcommand{\methodrow}{\rowcolor{TableHighlight}}
\newcolumntype{L}[1]{>{\raggedright\arraybackslash}p{#1}}
\newcolumntype{Y}{>{\centering\arraybackslash}X}

\newcolumntype{M}[1]{>{\raggedright\arraybackslash}m{#1}}

\definecolor{wacvblue}{rgb}{0.21,0.49,0.74}
\usepackage[breaklinks,colorlinks,allcolors=wacvblue]{hyperref}
\hypersetup{
  pdftitle={FracEvent: Event-Camera Simulation via Fractional-Relaxation Pixel Dynamics},
  pdfauthor={Langyi Chen; Chuanzhi Xu; Haoxian Zhou; Pengfei Ye; Ziyu Luo; Haodong Chen; Qiang Qu; Xiaoming Chen; Weidong Cai},
  pdfsubject={WACV 2027}
}

\def\wacvPaperID{502} 
\def\confName{WACV}
\def\confYear{2027}

\definecolor{MintDeep}{RGB}{10,130,130}
\definecolor{MintA}{RGB}{26,150,150}
\definecolor{MintB}{RGB}{46,168,168}
\definecolor{MintC}{RGB}{70,184,184}
\definecolor{MintD}{RGB}{94,198,198}

\title{
{\color{MintDeep}F}{\color{MintA}r}{\color{MintB}a}{\color{MintC}c}{\color{MintDeep}E}{\color{MintA}v}{\color{MintB}e}{\color{MintC}n}{\color{MintD}t}:
Event-Camera Simulation via Fractional-Relaxation Pixel Dynamics
}

\author{
Langyi Chen$^{1}$ \quad
Chuanzhi Xu$^{1}$\thanks{Corresponding author: chuanzhi.xu@sydney.edu.au} \quad
Haoxian Zhou$^{1}$ \quad
Pengfei Ye$^{2}$ \quad
Ziyu Luo$^{3}$ \\
Haodong Chen$^{1}$ \quad
Qiang Qu$^{1}$ \quad
Xiaoming Chen$^{3}$ \quad
Weidong Cai$^{1}$ \\
$^{1}$The University of Sydney, Australia \quad
$^{2}$Massachusetts Institute of Technology, United States \\
$^{3}$Beijing Technology and Business University, China
}

\newcommand{\maketeaserfigure}{%
  \vspace{-16pt}
  \noindent\begin{minipage}{\textwidth}
    \centering
    \includegraphics[width=\linewidth]{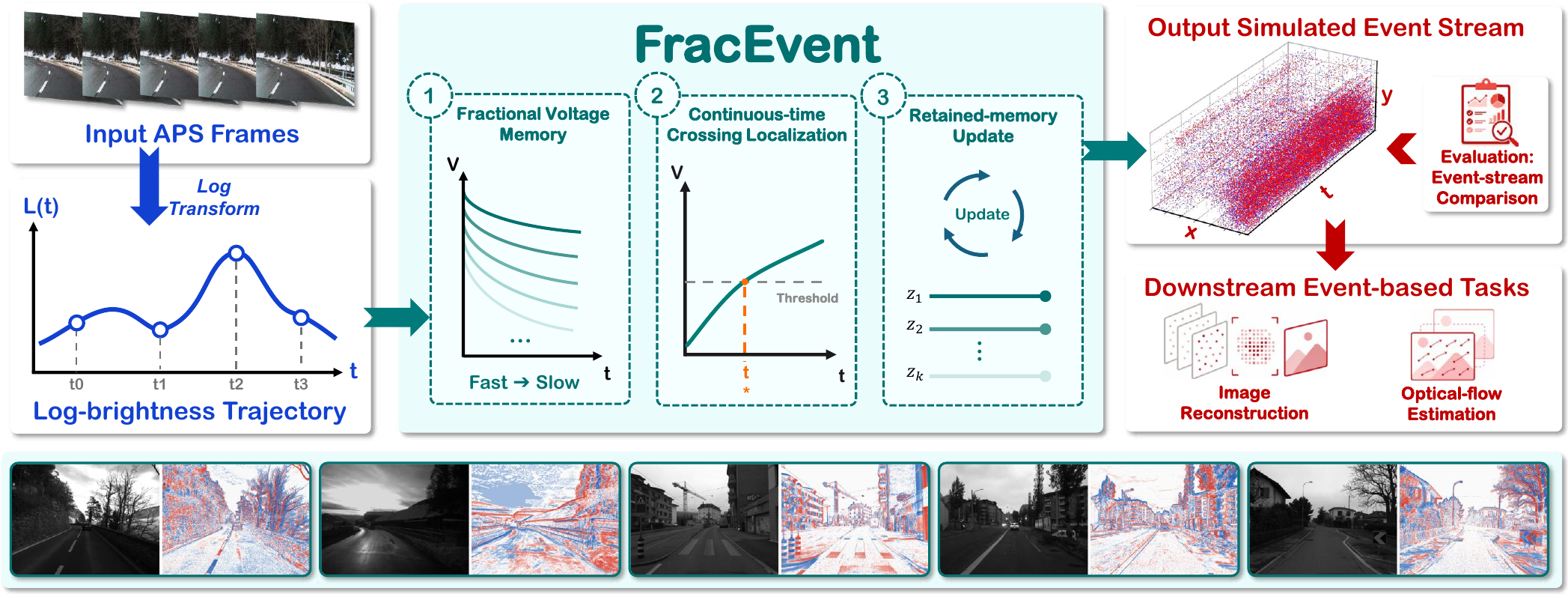}
    \captionsetup{hypcap=false}
    \captionof{figure}{\textbf{Overview.}
    Given APS frames, \method\ generates events through fractional voltage memory, continuous-time crossing localization, and retained-memory updates.
    We evaluate the resulting streams directly and through downstream tasks.}
    \label{fig:overview}
  \end{minipage}
  \vspace{12pt}
}

\makeatletter
\g@addto@macro\@maketitle{\maketeaserfigure}
\makeatother

\begin{document}
\raggedbottom
\maketitle
\newcommand{\abstractprojectpage}{~\href{https://bonaparte233.github.io/FracEvent/}{\textcolor{wacvblue}{Project page}}}

\providecommand{\abstractprojectpage}{}

\begin{abstract}
Event cameras asynchronously report brightness changes with microsecond-level temporal resolution, but real event data remain difficult to collect at scale because specialized sensors, careful synchronization, and task-specific annotations are required. Event-camera simulation is therefore important to event-based vision tasks. Most practical simulators build on contrast-threshold event generation, some with additional filtering, stochastic noise, or calibrated sensor parameters. While effective, such formulations can simplify the temporal structure produced by a pixel's recent response history. We introduce FracEvent, a physics-inspired sensor-side simulator with compact fractional-relaxation voltage dynamics. Given a log-intensity trajectory, FracEvent combines fast and slow response modes, emits ON/OFF events by localizing threshold crossings on the continuous voltage trajectory, and updates the event reference while retaining the mode states. This retained state links residual response to later event timing. We evaluate FracEvent through direct event-stream comparison and downstream transfer on image reconstruction and optical flow estimation. Across multiple datasets, FracEvent achieves leading results on multiple metrics in both evaluations, showing its practical value for event-camera simulation.\abstractprojectpage
\end{abstract}

\section{Introduction}
\label{sec:intro}

Event cameras, also known as dynamic vision sensors (DVS), report brightness changes as asynchronous pixel events.
At each pixel, the sensor emits a positive (ON) or negative (OFF) event when the accumulated log-brightness change crosses a contrast threshold.
This threshold-triggered readout provides microsecond-level temporal resolution and high dynamic range while concentrating output around temporal brightness changes \cite{lichtsteiner2008dvs,brandli2014davis}.
These properties make event sensing valuable for robotics, autonomous driving, optical flow estimation, and video reconstruction \cite{gallego2022eventsurvey,xu2025event3dsurvey}.
Representative systems use events for visual-inertial odometry, steering prediction, joint optical-flow/intensity estimation, and event-to-image reconstruction \cite{zhu2017evio,maqueda2018steering,bardow2016flowintensity,rebecq2019e2vid}.
However, collecting real event streams at scale requires dedicated event sensors and capture hardware, often with task-specific synchronization, annotation, and scene setups \cite{mvsec2018,dsec2021}.
Because event cameras remain relatively expensive and scarce, real event collection is hard to expand on demand across new scenes, motion conditions, and sensor assumptions \cite{ziegler2023realtime}.
At the same time, general computer vision research already provides large collections of high-quality frame-based video data.
Therefore, event-camera simulation serves as an important complement to real event data by turning frame-based data into event streams \cite{video2events2020,chen2026eventsim}.
However, this conversion is more than a simple change of data format.
Frame sequences and renderers provide per-pixel brightness trajectories, while the resulting events depend on when those trajectories drive pixel states across ON/OFF thresholds.
Their temporal sampling, interpolation, smoothing, and local contrast evolution can shift threshold-crossing times and change event counts \cite{rebecq2018esim,hu2020v2e,adv2e2024}.

Event simulation couples an \emph{input brightness trajectory}, supplied by rendering, video, frame interpolation, or raw-camera processing, with \emph{sensor-side conversion}, which determines event emission from each pixel's internal state.
Existing methods combine these choices in different ways, from continuous rendering \cite{rebecq2018esim,mujocoesim2023} or interpolation networks applied to input frames \cite{video2events2020}, to sensor-aware models of nonidealities and noise \cite{hu2020v2e,dvsvoltmeter2022,adv2e2024}, to learned and controllable generators \cite{eventgan2019,reliableflow2024,controlevents2026}.
During periods of high event activity, real streams often exhibit temporally sustained but irregular firing, including short-interval repetitions at the same pixel and asymmetric ON/OFF activity.
This structure motivates a sensor-side conversion model that retains recent response and lets it influence subsequent event timing.

Motivated by this observation, we propose \method, a physics-inspired event-camera simulator that uses fractional-relaxation dynamics to model sensor-side memory conditioned on a supplied log-intensity trajectory.
The input log-intensity drives several relaxation modes: fast modes follow immediate brightness changes, slow modes retain residual response, and their weighted sum forms a voltage trajectory.
Events are generated from continuous-time ON/OFF threshold crossings of this trajectory relative to a reference.
After each event, the reference moves while the relaxation modes remain active.
\Cref{fig:overview} summarizes this flow from input trajectory to simulated events and evaluation.

We evaluate \method\ against real events through quantitative statistics, qualitative comparisons, and ablations of temporal dynamics and event-generation components.
Downstream transfer uses E2VID-style reconstruction with established evaluation protocols \cite{rebecq2019e2vid,scheerlinck2020firenet,stoffregen2020simtoreal,evreal2023} and EV-FlowNet-style optical flow \cite{zhu2018evflownet,mvsec2018}.
To align calibration effort, we evaluate each baseline using both its public parameter setting and an event-only calibrated setting selected with the same candidate count used to calibrate \method\ for each sensor family.
Across these evaluations, \method\ gives the lowest reconstruction LPIPS on DAVIS240C and HQF, the lowest HQF MSE, and the lowest simulated-source optical flow mean AEE.

Our contributions can be summarized as follows:
\begin{itemize}
  \item We introduce \method, a physics-inspired event-camera simulator that models the \textbf{pixel event lifecycle} with \textbf{fractional-relaxation voltage memory}, preserving residual sensor state across repeated ON/OFF events.
  \item We formulate \textbf{trajectory-conditioned sensor-side conversion} for event simulation, allowing the same pixel dynamics to be paired with renderers, interpolation modules, frame sequences, or other sources that supply log-intensity trajectories.
  \item We develop a \textbf{continuous-time event-generation module} with closed-form mode updates, active-pixel bisection for triggered threshold crossings, event-wise reference updates, and separate ON/OFF thresholds.
  \item We validate \method\ through direct \textbf{event-stream comparison}, fixed-parameter ablations, and \textbf{downstream transfer}, reporting metric-specific gains and trade-offs under fixed protocols.
\end{itemize}

\section{Related Work}
\label{sec:related}

\noindent{\textbf{Input Trajectories for Event Simulation.}}
Event simulation begins with the brightness trajectory that drives each pixel.
Rendering-based simulators such as ESIM obtain this signal from continuous scene motion before applying an event rule, and later systems extend the same route to robotics, path tracing, contact-rich simulation, and task-specific synthetic data \cite{rebecq2018esim,mujocoesim2023,pecs2024,synsacc2026wacv,reinold2025wacvw}.
Video-to-event pipelines take a different route by reusing frame datasets.
In these pipelines, interpolation, exposure handling, filtering, and raw-camera processing shape the upstream trajectory \cite{video2events2020,hu2020v2e,v2ce2024,adv2e2024,raw2event2025}.
Missing motion, occlusion, or exposure changes in the input trajectory cannot be recovered by subsequent thresholding.
Given a supplied $L(t)$, \method\ isolates the complementary sensor-side conversion to events.

\noindent{\textbf{Sensor-Side Event Simulation.}}
Given a brightness trajectory, the remaining question is how each pixel converts it into asynchronous ON/OFF events.
Classical DVS and DAVIS models emit events when accumulated log-brightness change crosses a contrast threshold \cite{lichtsteiner2008dvs,brandli2014davis}, while practical simulators add interpolation, bandwidth limits, threshold mismatch, noise, and calibration \cite{hu2020v2e,rebecq2018esim,adv2e2024}.
The photoreceptor and source-follower response can be described by two real poles; v2e notes this second-order circuit response but adopts a dominant-pole first-order filter for practical simulation \cite{hu2020v2e}.
DVS-Voltmeter instead models stochastic internal-voltage drift and diffusion and samples events through a first-passage process, emphasizing photon/leakage noise and event-time variability \cite{dvsvoltmeter2022}.
Learned and domain-adaptive generators instead match event distributions or synthesize controllable event streams from data-driven priors \cite{eventgan2019,reliableflow2024,controlevents2026}.
\method\ follows the explicit sensor-side modeling direction by representing persistent response over a distribution of relaxation time scales and locating threshold crossings in continuous time.
Unlike stochastic drift/diffusion, the deterministic state trajectory isolates history-dependent timing and is repeatable for fixed inputs and parameters.

\noindent{\textbf{Downstream Applications of Simulated Event Streams.}}
Simulated events also support downstream event-based vision tasks, whose representations and temporal structure are central to 3D reconstruction, event-stream super-resolution, and pose estimation \cite{xu2025voxel3d,xu2026eventsr,zhou2025eventpose}.
Here we focus on image reconstruction and optical flow because they provide mature and complementary evaluations: reconstruction tests whether accumulated events preserve recoverable appearance, while flow tests whether event timing and local motion structure support motion estimation.
For image reconstruction, E2VID- and FireNet-style recurrent models map event tensors to intensity images \cite{rebecq2019e2vid,scheerlinck2020firenet}, with later work systematizing reconstruction protocols and sim-to-real reporting \cite{fox2024wacv,stoffregen2020simtoreal,evreal2023}.
For optical flow, EV-FlowNet learns dense flow from events with self-supervised photometric losses and evaluates on MVSEC \cite{zhu2018evflownet,mvsec2018}; later work adds stronger architectures, higher-resolution driving datasets, and contrast-maximization objectives \cite{gehrig2021eraft,dsec2021,karmokar2025wacv}.
We therefore evaluate both tasks with the network, representation, schedule, and test protocol fixed while only the simulated event source changes.

\section{Methodology}
\label{sec:method}

\begin{figure*}[t]
  \centering
  \includegraphics[width=\textwidth]{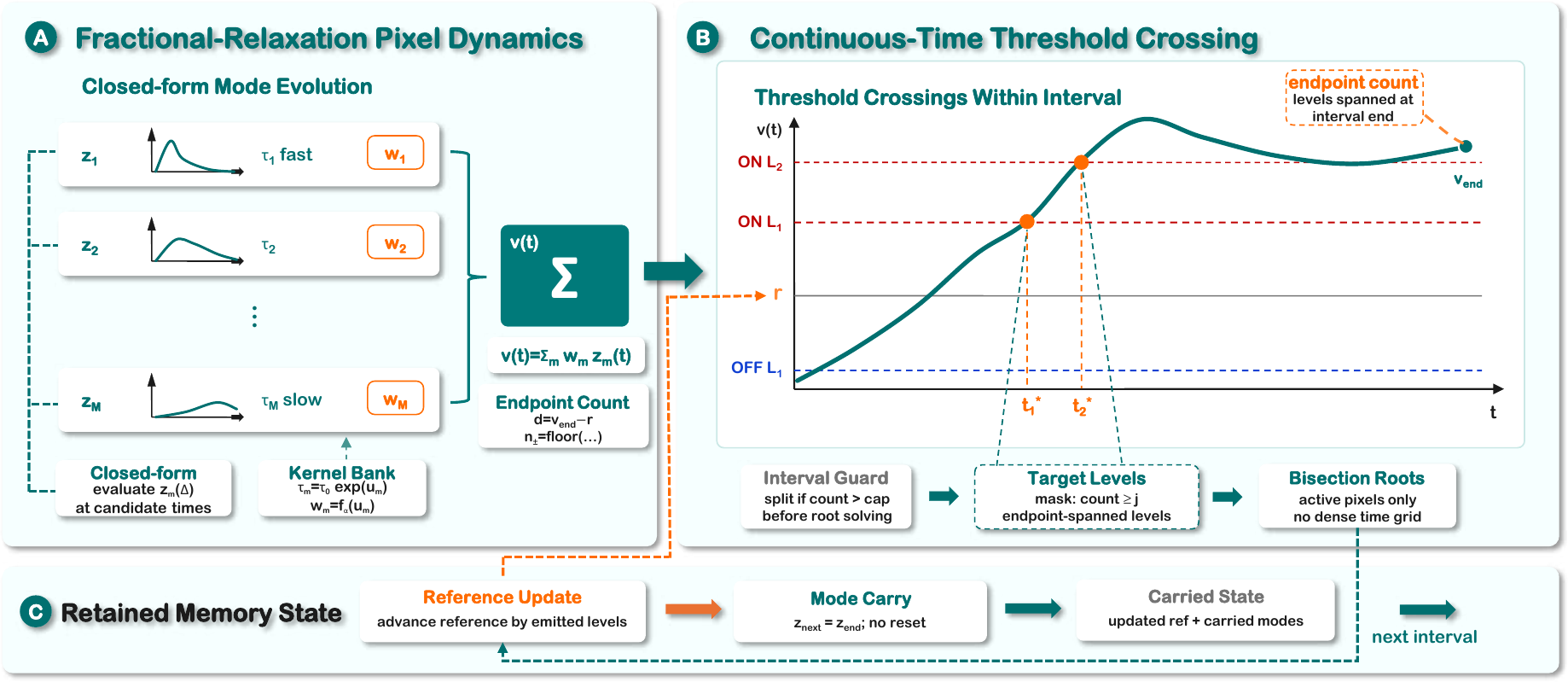}
\caption{\textbf{Equivalent pixel-dynamics view of \method.}
  Fractional-memory modes convert the log-intensity trajectory into a voltage state.
  Events arise from continuous-time ON/OFF crossings relative to a reference updated after each event, while the modes are retained.}
  \label{fig:method}
\end{figure*}

\subsection{Problem Setup for Event-Stream Simulation}
\label{sec:problem}

At each pixel, let $I(t)$ denote a normalized intensity trajectory supplied by an upstream source, such as a frame sequence, interpolation module, or renderer.
We use the log-intensity form common in contrast-threshold event-camera models:
\begin{equation}
  L(t)=\log(I(t)+i_{\mathrm{dark}}+\epsilon).
  \label{eq:log_intensity}
\end{equation}
Here $i_{\mathrm{dark}}$ and $\epsilon$ are small positive offsets that stabilize the log transform near zero intensity.
We adopt log intensity as the conventional relative-brightness representation for contrast-threshold event models and use the resulting scalar trajectory to drive the relaxation dynamics.
The relaxation equations can use another scalar photoreceptor representation as the input drive within the same formulation.
Given $L(t)$, \method\ generates an asynchronous event stream:
\begin{equation}
  \events=\{(x_i,y_i,t_i,p_i)\}_{i=1}^{N},
\end{equation}
where $(x_i,y_i)$ is the pixel coordinate, $t_i$ is the event timestamp, and $p_i\in\{+1,-1\}$ is the event polarity, with $p_i=+1$ denoting an ON event and $p_i=-1$ denoting an OFF event.

Fixing $L(t)$ separates the upstream trajectory from the pixel rule that converts brightness changes into events.
This lets the same sensor-side model be paired with different upstream sources while isolating how pixel dynamics affect timing, polarity, and density.
Real event streams often contain repeated same-pixel triggers and short bursts, so the pixel rule must carry residual state beyond the current frame-to-frame contrast.
We instantiate this rule with fractional-relaxation voltage memory, continuous-time threshold crossing, and reference updates that retain the memory state. \Cref{fig:method} shows how these components form this sensor-side conversion model.

\subsection{Fractional-Relaxation Pixel Dynamics}
\label{sec:frac_modes}

\method\ represents recent response history with fast and slow relaxation channels.
Fast channels follow immediate changes, while slow channels retain residual response across later reference updates.
Only their weighted voltage is compared with the event reference; the individual channels are carried forward as memory states.
The channels are placed at dimensionless log-time coordinates:
\begin{equation}
  \xi_m \in [-R,R], \qquad m=1,\ldots,M,
  \label{eq:mode_offsets}
\end{equation}
sampled uniformly over the configured relaxation log range $R$.
Each coordinate $\xi_m$ selects one relaxation time scale in the quadrature.
For a fractional shape parameter $\alpha\in(0,1)$, the mode weights are:
\begin{equation}
  \tilde{w}_m =
  \frac{\sin(\pi\alpha)}{\cosh(\alpha \xi_m)+\cos(\pi\alpha)}, \qquad
  w_m=\frac{\tilde{w}_m}{\sum_j \tilde{w}_j},
  \label{eq:fractional_weights}
\end{equation}
where the denominator normalizes the weights over all modes.
The first-order control uses the single-mode definition $M=1$ and $w_1=1$.

For mode $m$, let $z_m(t)$ be a memory channel with time constant $\tau_m$.
Each channel follows a first-order relaxation equation:
\begin{equation}
  \frac{d z_m(t)}{dt}=\frac{L(t)-z_m(t)}{\tau_m}.
  \label{eq:mode_ode}
\end{equation}
The pixel voltage, denoted by $v(t)$, is the weighted sum of these channels:
\begin{equation}
  v(t)=\sum_{m=1}^{M} w_m z_m(t).
  \label{eq:voltage_sum}
\end{equation}
Together, the weighted channels form a compact finite-mode approximation to a broad fractional-memory kernel.
This distributed-exponential construction follows standard fractional-relaxation representations and the classical distributed-relaxation view behind Cole-Cole dispersion \cite{cole1941dispersion,podlubny1999fractional,mainardi2010fractional}.
Fast modes track immediate brightness changes, while slow modes preserve residual response over longer intervals.
The physical motivation is finite-bandwidth response before thresholding: photoreceptor and subsequent pixel stages introduce light-dependent relaxation and latency \cite{joubert2021characterized}.
At fixed brightness, a first-order response has one relaxation scale and a two-pole response has two; a distributed kernel permits a broader residual-response tail within the same state-update framework.
We use the Cole--Cole distribution as a compact phenomenological parameterization of this distributed response and compare its empirical behavior with first-order and second-order dynamics.
For each frame interval, the mode time constants are set from the mid-interval brightness:
\begin{equation}
  \tau_0 =
    \tau_{\mathrm{ref}}
    \left(\frac{I_{\mathrm{ref}}}{\bar I+i_{\mathrm{dark}}}\right)^\beta,
  \qquad
  \tau_m = \tau_0 e^{\xi_m},
  \label{eq:brightness_tau}
\end{equation}
where $\bar I=(I(t_k)+I(t_{k+1}))/2$, $\tau_{\mathrm{ref}}$ and $I_{\mathrm{ref}}$ set the response scale and reference brightness, and $\beta$ controls the brightness dependence.
We bound $\tau_0$ to $[\tau_{\min},\tau_{\max}]$ to avoid unrealistically slow or fast responses.
The uniform log-time grid yields geometrically spaced time constants, allowing a small number of modes to cover short and long response scales.
At sequence start, all modes are initialized from the first log-intensity frame, and the reference is initialized as their weighted sum.

After $\tau_m$ is fixed for an interval, each memory channel can be evaluated in continuous time.
We assume a linear log-intensity segment within the interval:
\begin{equation}
  L(t_k+\Delta)=L_0+s\Delta,
  \qquad
  0\leq\Delta\leq dt.
  \label{eq:linear_segment}
\end{equation}
Here $dt=t_{k+1}-t_k$, $L_0=L(t_k)$, $s=(L(t_{k+1})-L_0)/dt$, and $a_m(\Delta)=e^{-\Delta/\tau_m}$.
Each mode then has the closed-form update:
\begin{equation}
\begin{split}
z_m(t_k+\Delta)=&
a_m(\Delta)z_m(t_k)
 +(1-a_m(\Delta))L_0\\
&+s\left[\Delta-\tau_m(1-a_m(\Delta))\right].
\end{split}
\label{eq:closed_form}
\end{equation}
This expression gives the voltage state at any candidate time in the interval.
The threshold solver below uses this continuous-time state to assign event timestamps without a dense temporal grid.
\begin{table*}[t]
  \centering
  \caption{\textbf{Evaluation setup for event-stream comparison and downstream tasks.}}
  \label{tab:datasets}
  \footnotesize
  \setlength{\tabcolsep}{3.5pt}
  \renewcommand{\arraystretch}{1.05}
  \begin{tabular}{@{}>{\raggedright\arraybackslash}m{0.16\textwidth}>{\raggedright\arraybackslash}m{0.21\textwidth}>{\raggedright\arraybackslash}m{0.42\textwidth}>{\raggedright\arraybackslash}m{0.16\textwidth}@{}}
    \toprule
    Evaluation & Dataset & Protocol & Metrics \\
    \midrule
    \rowcolor{EvalStream}
    Event-stream comparison & DAVIS240C windows \cite{mueggler2017dataset} and DAVIS346 real-life captures & Same frame/timestamp windows and metrics, with simulated streams differing only in sensor-side dynamics. & Count, IEI, polarity, time surface, spatial, voxel, repeated IEI \\
    \rowcolor{EvalRecon}
    Image reconstruction & GoPro train \cite{nah2017gopro} and DAVIS240C/HQF tests \cite{mueggler2017dataset,stoffregen2020simtoreal} & Same resolution, E2VID-style model, loss, optimizer, schedule, and test sets. & MSE, SSIM, LPIPS \\
    \rowcolor{EvalFlow}
    Optical flow estimation & MVSEC Day2 train and Day1/indoor tests \cite{mvsec2018} & EV-FlowNet-style network, event representation, losses, 300k schedule, and evaluation splits. & AEE, outlier percentage \\
    \bottomrule
  \end{tabular}
\end{table*}

\subsection{Continuous-Time Threshold Crossing}
\label{sec:crossing}

\method\ emits events when the voltage state crosses threshold levels relative to the reference.
Let $r$ denote the reference voltage carried into the current interval, and define the reference-relative voltage:
\begin{equation}
  d(t)=v(t)-r,
  \label{eq:relative_voltage}
\end{equation}
An ON event is generated when $d(t)\geq\thetaon$, and an OFF event is generated when $d(t)\leq-\thetaoff$, where $\thetaoff$ is stored as a positive magnitude.

For each frame interval, the solver first computes ON/OFF event counts from the endpoint reference-relative voltage:
\begin{equation}
  n^+ = \left\lfloor \frac{[d(t_{k+1})]_+}{\thetaon}\right\rfloor,\qquad
  n^- = \left\lfloor \frac{[-d(t_{k+1})]_+}{\thetaoff}\right\rfloor.
  \label{eq:event_counts}
\end{equation}
Here $[q]_+=\max(q,0)$ denotes the positive part.
These counts define the ON/OFF threshold levels relative to the interval-start reference.
For each active pixel and each ON level $j=1,\ldots,n^+$, it solves
$v(t)=r+j\thetaon$ by bisection on $[t_k,t_{k+1}]$.
For each OFF level $j=1,\ldots,n^-$, it solves $v(t)=r-j\thetaoff$ in the same way.
The bisection stage assigns continuous-time timestamps to these counted levels for active pixels, and the resulting events are sorted by timestamp before output.
This is a local root-localization step: on monotone pixel subintervals, each counted level has a unique crossing.
If the endpoint count in an interval exceeds the solver cap, the interval is recursively split before bisection, so dense bursts are handled without a uniform substep grid.

After emitting these levels, \method\ advances only the reference state:
\begin{equation}
  r\leftarrow r+\thetaon\;\text{(ON)},\qquad r\leftarrow r-\thetaoff\;\text{(OFF)}.
\end{equation}
For multiple same-polarity levels in one interval, the same step is applied once per emitted level.
The mode states $z_m(t)$ are not reset.
Thus, an event shifts the reference voltage while leaving residual voltage memory available for the next crossing.

This lifecycle gives the simulated stream three useful properties.
First, preserving the fractional modes carries sub-threshold response across events, which supports repeated triggering and inter-event-interval structure.
Second, continuous-time bisection assigns timestamps to counted threshold levels inside each frame interval, producing asynchronous event times and avoiding grid-locked emissions.
Third, separate ON/OFF reference updates represent polarity asymmetry with explicit threshold magnitudes.
Together, these properties make density, timing, and polarity reflect sensor-side dynamics conditioned on the supplied input trajectory.

\Cref{alg:core_algorithm} gives the core sensor-side conversion loop; the corresponding implementation and additional details are provided in the supplementary code and Supplementary Material.

\begin{algorithm}[h]
  \caption{Sensor-side event generation in \method}
  \label{alg:core_algorithm}
  \begin{algorithmic}[1]
    \Statex \textbf{Input:} frames $\{I(t_k)\}_{k=0}^{K}$, timestamps, parameters
    \Statex \textbf{Output:} sorted event stream $\events$
    \State Convert $I(t_k)$ to log intensity $L(t_k)$ by \Cref{eq:log_intensity}
    \State Initialize $z_m \gets L(t_0)$ for all modes and $\events\gets\emptyset$
    \State Set $r\gets v(t_0)$ by \Cref{eq:voltage_sum}
    \For{$k=0,\ldots,K-1$}
      \State Set interval time constants $\tau_m$ by \Cref{eq:brightness_tau}
      \State Define $L(t)$ on $[t_k,t_{k+1}]$ by \Cref{eq:linear_segment}
      \State Evaluate endpoint voltage $v(t_{k+1})$ with \Cref{eq:closed_form,eq:voltage_sum}
      \State Store $r_0\gets r$
      \State Compute endpoint counts $n^+,n^-$ using \Cref{eq:event_counts}
      \If{$\max(n^+,n^-)$ exceeds the solver cap}
        \State Recursively process the two split subintervals
        \State \textbf{continue}
      \EndIf
      \For{each active pixel and counted threshold level}
        \State Locate the event timestamp $t^\star$ by bisection
        \State Append $(x,y,t^\star,p)$ to $\events$
      \EndFor
      \State Update $r\gets r_0+n^+\thetaon-n^-\thetaoff$ pixelwise
      \State Carry the mode states $z_m(t_{k+1})$ to the next interval
    \EndFor
    \State Sort $\events$ by timestamp and return it
  \end{algorithmic}
\end{algorithm}

\section{Experiments and Results}
\label{sec:experiments}

\subsection{Experimental Setup}
\label{sec:setup}

\begin{figure}[!htbp]
  \centering
  \includegraphics[width=\linewidth]{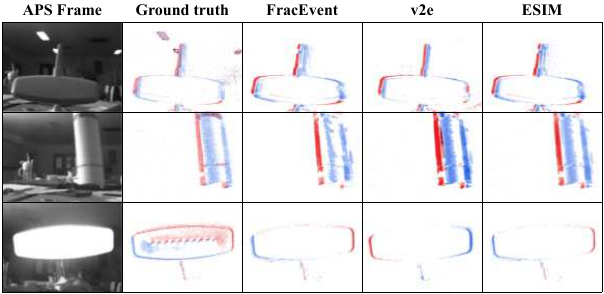}
  \caption{\textbf{DAVIS346 real-camera comparison in real-life scenes.}
  APS frames, real events, and events simulated by \method, v2e, and ESIM over matched frame intervals.}
  \label{fig:real_davis346_visual}
\end{figure}

We evaluate \method\ through event-stream comparisons over matched timestamp windows, including qualitative visualizations, and downstream reconstruction and optical flow against baseline simulators under a fixed protocol.
Each downstream configuration is trained once.
\Cref{tab:datasets} summarizes the datasets, protocols, and metrics.
We use events and APS frames from the DAVIS240C dataset for event-stream comparison and reconstruction evaluation \cite{brandli2014davis,mueggler2017dataset}.
GoPro high-frame-rate video supplies input for matched simulated event generation \cite{nah2017gopro}, HQF serves as an additional image reconstruction test set \cite{stoffregen2020simtoreal}, and MVSEC contains event streams, APS frames, and ground-truth flow for optical flow estimation training and evaluation \cite{mvsec2018}.
\Cref{fig:real_davis346_visual} shows an additional qualitative DAVIS346 check outside the quantitative DAVIS240C aggregate.

For simulator baselines, we compare ESIM \cite{rebecq2018esim}, v2e \cite{hu2020v2e}, and DVS-Voltmeter \cite{dvsvoltmeter2022} using public and calibrated parameter settings.
Public (P) settings come from the original papers or official repositories; calibrated (C) settings are selected using event-stream statistics.
We calibrate \method\ for each sensor family using real-event statistics, evaluating 240 candidates for DAVIS240 and 588 for DAVIS346.
For the reported comparisons, each baseline is likewise selected from a fixed candidate grid on the same sensor-family data and event-only objective, using the same candidate budget for the corresponding sensor family.
For DAVIS240, the objective weights count, temporal-histogram, polarity, and time-surface penalties by $1.0/0.5/0.35/0.25$; for DAVIS346, it combines anchor and context scores with weights $0.8/0.2$ and adds a unit-weight anchor-density term.
The DAVIS240 calibration sequences are disjoint from the four full-duration sequences used for event-stream comparison.
For DAVIS346, the 20 anchor windows come from \texttt{outdoor\_day2}, and eight unlabeled context windows come from the four evaluation sequences.
Parameter selection uses only event-stream statistics; downstream losses, test annotations, networks, and task-evaluation metrics are excluded, and the selected settings are fixed before training-stream generation.
The DAVIS240 event-stream comparison uses the same selected 240-candidate settings and disjoint held-out windows.
Because the downstream datasets use DAVIS240- and DAVIS346-series sensors, we use separate settings for the two sensor families.
Equal candidate counts align calibration data, objectives, and evaluation counts, while simulator-specific parameter spaces retain their native dimensions and ranges; the objectives, search ranges, and selected settings are given in Supplementary Material Secs. A--C.
The calibrated v2e control searches thresholds with its intensity-dependent cutoff and optional noise effects disabled, and ESIM uses zero refractory period. Runtime and memory measurements are reported in Supplementary Material Sec.~G.

We first compare simulated event streams with real DAVIS240C events over 256 deterministic 50 ms windows spread across the full APS durations of four held-out sequences.
For each simulated/real pair, Count measures density mismatch as the absolute log count-ratio error; IEI distance compares the distributions of logarithmic same-pixel intervals by 1-Wasserstein distance \cite{dvsvoltmeter2022,villani2009optimaltransport}; Polarity compares ON-event fractions; and Time Surface measures recent spatial-polarity agreement \cite{lagorce2017hots,sironi2018hats}.
The supplementary controls additionally report normalized per-pixel Spatial error, normalized polarity-separated Voxel error, and Repeated IEI error, defined from the fraction of same-pixel intervals no longer than 10 ms. IEI and Repeated IEI use the same 255 valid paired windows; complete definitions and sequence-cluster intervals are provided in the supplement.

\subsection{Event-Stream Comparison}
\label{sec:event_stream_comparison}

\begin{table}[t]
  \centering
  \caption{\textbf{Full-sequence event-stream results on DAVIS240C.}
  Mean errors over 256 held-out windows (255 for IEI); C/P denote 240-candidate calibrated/public settings. Lower is better.}
  \label{tab:event_stream_baselines}
  \scriptsize
  \setlength{\tabcolsep}{2.0pt}
  \renewcommand{\arraystretch}{0.96}
  \begin{tabularx}{\linewidth}{@{}lYYYY@{}}
    \toprule
    Method & Count $\downarrow$ & IEI $\downarrow$ & Polarity $\downarrow$ & Time surf. $\downarrow$ \\
    \midrule
    \methodrow \method\ (C=P) & 0.6535 & \best{0.0978} & \best{0.0303} & 0.6284 \\
    DVS-Voltmeter (C) & 0.6020 & 0.1915 & 0.0811 & \best{0.5609} \\
    DVS-Voltmeter (P) & 0.6912 & 0.2281 & 0.1458 & 0.5919 \\
    v2e (C) & \best{0.5341} & 0.2741 & 0.0471 & 0.5970 \\
    v2e (P) & 1.0793 & 0.4460 & 0.0816 & 0.7425 \\
    ESIM (C) & 0.6649 & 0.2176 & 0.0641 & 0.6462 \\
    ESIM (P) & 1.6635 & 0.6151 & 0.0790 & 0.5983 \\
    \bottomrule
  \end{tabularx}
\end{table}

\begin{figure}[!htbp]
  \centering
  \includegraphics[width=\linewidth]{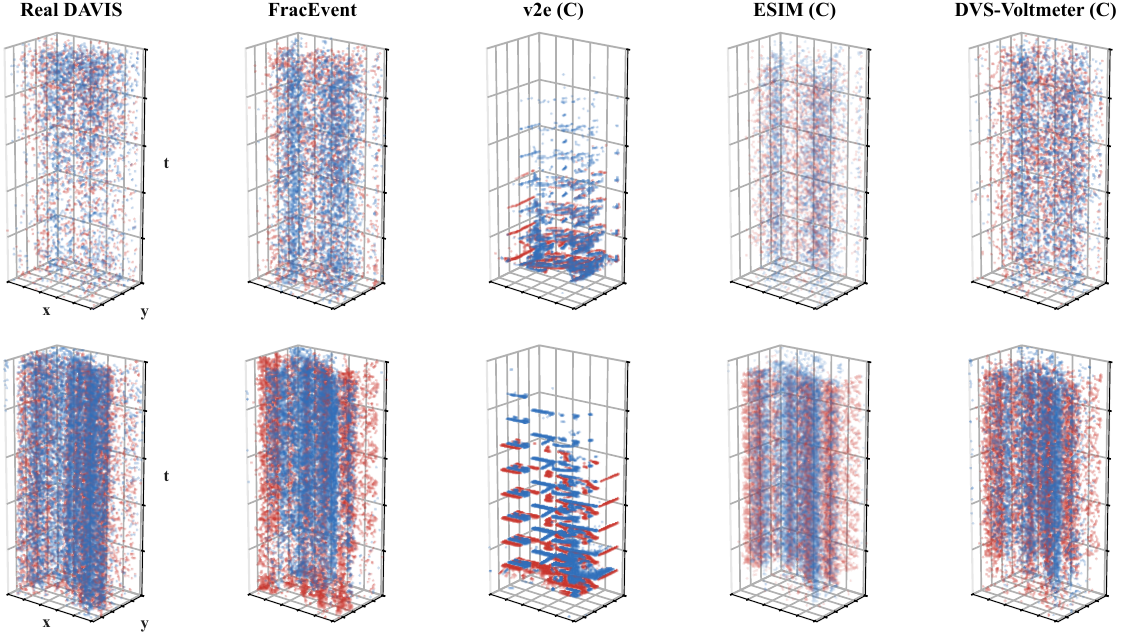}
  \caption{\textbf{Qualitative event-stream comparison.}
  Spatiotemporal plots over matched 50 ms DAVIS240C windows. C denotes 240-candidate calibrated settings; red/blue indicate ON/OFF events.}\label{fig:event_stream_mechanism_contrast}
\end{figure}
\begin{table*}[t]
  \centering
  \caption{\textbf{Optical flow estimation results on MVSEC.}
  P/C denote public and 588-candidate calibrated settings; \method\ has P=C. Real MVSEC is the real-event control; bold marks the best simulated-source AEE in each column.}
  \label{tab:mvsec}
  \footnotesize
  \setlength{\tabcolsep}{1.8pt}
  \renewcommand{\arraystretch}{0.96}
  \begin{adjustbox}{max width=\textwidth}
  \begin{tabular}{@{}lrrrrrrrrrr!{\hspace{3pt}\vrule\hspace{3pt}}rrrrrrrrrr@{}}
    \toprule
    & \multicolumn{10}{c}{$dt=1$ frame} & \multicolumn{10}{c}{$dt=4$ frames} \\
    \cmidrule(lr){2-11}\cmidrule(lr){12-21}
    & \multicolumn{2}{c}{Outdoor1} & \multicolumn{2}{c}{Indoor1} & \multicolumn{2}{c}{Indoor2} & \multicolumn{2}{c}{Indoor3} & \multicolumn{2}{c}{Mean}
    & \multicolumn{2}{c}{Outdoor1} & \multicolumn{2}{c}{Indoor1} & \multicolumn{2}{c}{Indoor2} & \multicolumn{2}{c}{Indoor3} & \multicolumn{2}{c}{Mean} \\
    Method & AEE & Out. & AEE & Out. & AEE & Out. & AEE & Out. & AEE & Out.
    & AEE & Out. & AEE & Out. & AEE & Out. & AEE & Out. & AEE & Out. \\
    \midrule
    Real MVSEC & 0.55 & 0.4 & 1.03 & 2.2 & 1.60 & 13.4 & 1.48 & 11.3 & 1.17 & 6.8
    & 1.53 & 12.0 & 3.14 & 37.1 & 5.49 & 53.5 & 4.52 & 47.8 & 3.67 & 37.6 \\
    \cmidrule(lr){1-21}
    \methodrow \method\ (P=C) & 0.63 & 0.3 & \best{1.07} & 2.2 & \best{1.71} & 14.8 & \best{1.57} & 13.7 & \best{1.24} & 7.8
    & 1.91 & 15.3 & \best{3.39} & 41.1 & \best{5.85} & 53.0 & \best{5.34} & 50.4 & \best{4.12} & 40.0 \\
    DVS-Voltmeter (P) & \best{0.59} & 0.3 & 1.12 & 2.1 & 1.83 & 16.0 & 1.66 & 12.4 & 1.30 & 7.7
    & \best{1.85} & 17.7 & 3.87 & 48.7 & 6.44 & 65.4 & 5.62 & 61.2 & 4.44 & 48.3 \\
    DVS-Voltmeter (C) & 0.62 & 0.3 & 1.10 & 1.9 & 1.80 & 15.4 & 1.63 & 12.3 & 1.29 & 7.5
    & 1.92 & 17.8 & 3.65 & 46.4 & 6.25 & 62.5 & 5.48 & 58.8 & 4.33 & 46.4 \\
    ESIM (P) & 0.65 & 0.4 & 1.26 & 4.6 & 2.13 & 22.7 & 1.90 & 17.3 & 1.49 & 11.2
    & 2.19 & 22.8 & 4.76 & 64.6 & 7.50 & 74.9 & 6.67 & 73.5 & 5.28 & 58.9 \\
    ESIM (C) & 1.07 & 1.2 & 1.35 & 7.6 & 2.10 & 20.9 & 2.00 & 20.9 & 1.63 & 12.7
    & 2.69 & 30.5 & 5.20 & 49.4 & 9.67 & 59.1 & 8.31 & 54.8 & 6.47 & 48.5 \\
    v2e (P) & 0.76 & 0.5 & 1.27 & 4.7 & 2.14 & 22.7 & 1.93 & 18.0 & 1.53 & 11.5
    & 2.53 & 29.2 & 4.62 & 61.1 & 7.31 & 74.4 & 6.54 & 73.5 & 5.25 & 59.6 \\
    v2e (C) & 0.73 & 0.3 & 1.12 & 2.9 & 1.87 & 18.3 & 1.70 & 16.1 & 1.35 & 9.4
    & 2.31 & 22.8 & 3.79 & 47.7 & 6.41 & 60.9 & 5.56 & 56.9 & 4.52 & 47.1 \\
    \bottomrule
  \end{tabular}
  \end{adjustbox}
\end{table*}

\begin{figure*}[!t]
  \centering
  \includegraphics[width=\textwidth]{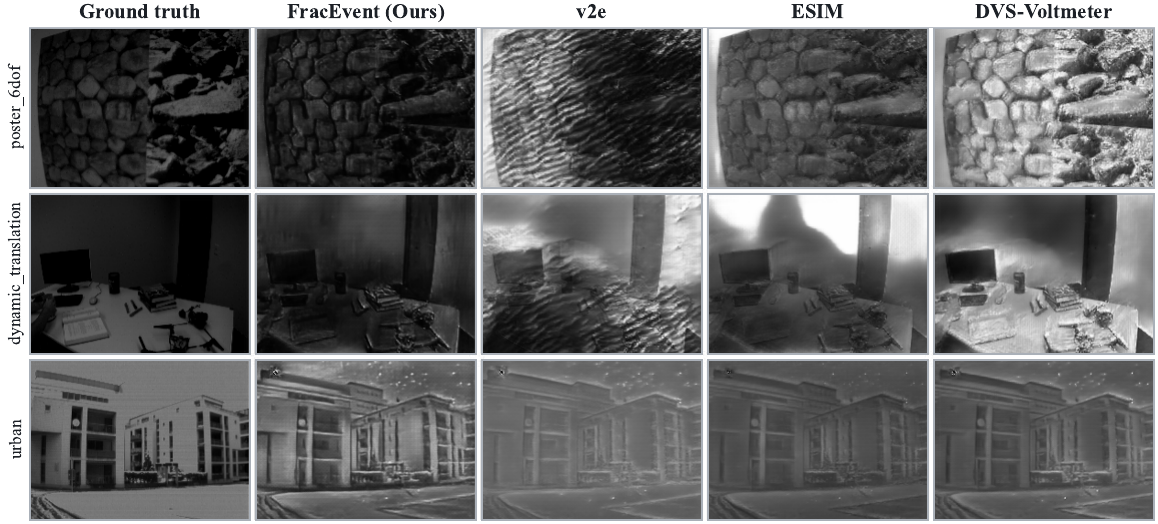}
  \captionsetup{justification=raggedright,singlelinecheck=false}
  \caption{\textbf{Image reconstructions on DAVIS240C.}
  Representative examples from models trained with different simulated event streams. Raw $[0,1]$ outputs are shown without per-method contrast normalization.}
  \label{fig:reconstruction}
\end{figure*}

\noindent\textbf{Baseline Comparison.}
\Cref{tab:event_stream_baselines} reports DAVIS240C event-stream errors under public and 240-candidate calibrated settings.
\method\ has the lowest IEI and Polarity errors, calibrated v2e has the lowest Count error, and calibrated DVS-Voltmeter has the lowest Time Surface error.
Calibration materially changes ESIM, v2e, and DVS-Voltmeter, but not uniformly: it improves several density/timing measures while worsening selected Polarity, Voxel, or Repeated IEI measures.
The calibrated-setting visualization in \Cref{fig:event_stream_mechanism_contrast} qualitatively illustrates the complementary density, timing, polarity, and recency behavior summarized by the metrics.

\noindent\textbf{Component Ablation.}
Under the same window protocol, each ablation removes one component of \method\ while keeping the input frames, parameter settings, and metrics fixed.
The tested variants remove multi-mode memory, retained memory state, or separate ON/OFF thresholds.

{
\begin{table}[!htbp]
  \centering
  \caption{\textbf{\method\ component ablation.}
  Mean held-out event-stream errors; lower is better.}
  \label{tab:component_comparison}
  \footnotesize
  \setlength{\tabcolsep}{2.2pt}
  \renewcommand{\arraystretch}{1}
  \begin{adjustbox}{max width=\linewidth}
  \begin{tabular}{@{}lcccc@{}}
    \toprule
    Variant & Count & IEI & Polarity & Time surf. \\
    \midrule
    \methodrow Full \method & 0.6535 & \best{0.0978} & \best{0.0303} & 0.6284 \\
    w/o multi-mode memory & \best{0.6256} & 0.1010 & 0.0475 & \best{0.6124} \\
    w/o retained memory state & 0.6566 & 0.1047 & 0.0492 & 0.6298 \\
    w/o separate thresholds & 0.6598 & 0.1070 & 0.0987 & 0.6285 \\
    \bottomrule
  \end{tabular}
  \end{adjustbox}
\end{table}
}

\begin{figure*}[!t]
  \centering
  \includegraphics[width=\textwidth]{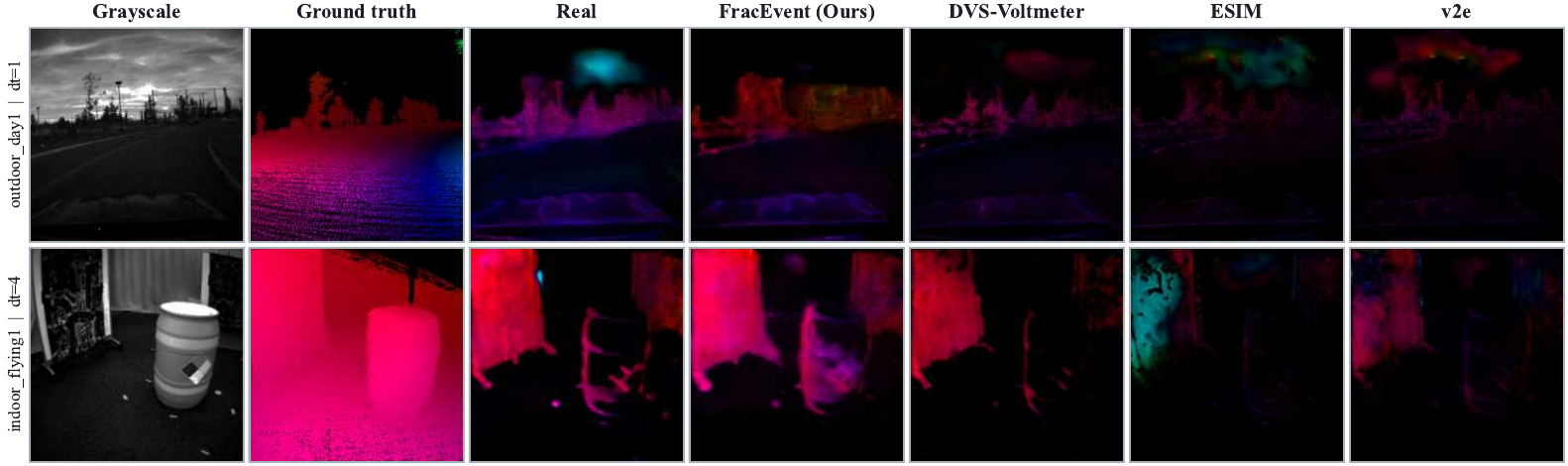}
  \captionsetup{justification=raggedright,singlelinecheck=false}
  \caption{\textbf{Optical-flow predictions on MVSEC.}
  Representative outdoor and indoor examples.}
  \label{fig:flow_qualitative}
\end{figure*}

\Cref{tab:component_comparison} shows metric-specific effects of cross-interval state retention and separate ON/OFF thresholds.
Multi-mode memory improves IEI and Polarity errors, while the single-mode control performs better on Count and Time Surface.
Supplementary Material Sec.~D provides the complete metric definitions, all seven reported metrics, and the fixed-parameter temporal-dynamics comparison with first-order, second-order, and Uniform-4 controls.

\begin{table}[!htbp]
  \centering
  \caption{\textbf{Image reconstruction results on DAVIS240C and HQF.}
  P/C denote public and 240-candidate calibrated settings; \method\ has P=C. Bold marks the best result in each column.}
  \label{tab:reconstruction}
  \footnotesize
  \setlength{\tabcolsep}{2.2pt}
  \renewcommand{\arraystretch}{0.96}
  \begin{adjustbox}{max width=\linewidth}
  \begin{tabular}{@{}lrrrrrr@{}}
    \toprule
    & \multicolumn{3}{c}{DAVIS240C} & \multicolumn{3}{c}{HQF} \\
    \cmidrule(lr){2-4}\cmidrule(lr){5-7}
    Train source & MSE $\downarrow$ & SSIM $\uparrow$ & LPIPS $\downarrow$ & MSE $\downarrow$ & SSIM $\uparrow$ & LPIPS $\downarrow$ \\
    \midrule
    \methodrow \method\ (P=C) & 0.030 & 0.371 & \best{0.460} & \best{0.052} & 0.355 & \best{0.443} \\
    v2e (P) & 0.065 & 0.253 & 0.575 & 0.073 & 0.304 & 0.556 \\
    v2e (C) & 0.030 & 0.384 & 0.499 & 0.095 & 0.307 & 0.552 \\
    ESIM (P) & 0.054 & \best{0.411} & 0.496 & 0.080 & 0.372 & 0.551 \\
    ESIM (C) & 0.034 & 0.281 & 0.479 & 0.076 & 0.326 & 0.445 \\
    DVS-Voltmeter (P) & 0.129 & 0.275 & 0.561 & 0.086 & 0.351 & 0.508 \\
    DVS-Voltmeter (C) & \best{0.027} & 0.397 & 0.463 & 0.072 & \best{0.383} & 0.482 \\
    \bottomrule
  \end{tabular}
  \end{adjustbox}
\end{table}

\subsection{Image Reconstruction}
\label{sec:reconstruction}

For each simulated source, we train the same E2VID-style recurrent model for 20 epochs on the full GoPro split at 240 fps and $240\times180$ resolution \cite{rebecq2019e2vid,nah2017gopro}; only the event source changes, and valid-window differences yield 444,000--452,620 updates.
We evaluate full DAVIS240C and HQF sequences \cite{stoffregen2020simtoreal} using MSE, SSIM \cite{wang2004ssim}, and LPIPS \cite{zhang2018lpips}, following EVREAL reporting \cite{evreal2023}.

\Cref{tab:reconstruction} compares public (P) and 240-candidate calibrated (C) settings under the same downstream protocol.
Across all reported settings, \method\ leads three of the six metrics: DAVIS240C LPIPS and HQF MSE/LPIPS. DVS-Voltmeter (C) leads DAVIS240C MSE and HQF SSIM, while ESIM (P) leads DAVIS240C SSIM.
These results measure the utility of simulated training data for reconstruction against the chosen APS targets.
\Cref{fig:reconstruction} presents representative reconstructions from models trained with different simulated event sources.

\subsection{Optical Flow Estimation}
\label{sec:flow}

For each source, we train the same EV-FlowNet-style self-supervised model on MVSEC \texttt{outdoor\_day2} for 300k steps \cite{zhu2018evflownet}, fixing model, loss, schedule, and evaluation code; Real MVSEC is a separately trained reference.
We evaluate \texttt{outdoor\_day1} and \texttt{indoor\_flying1/2/3} at $dt=1,4$, reporting AEE and event-masked KITTI-style outliers (error above 3 px and 5\% of flow magnitude).

As shown in \Cref{tab:mvsec}, \method\ has the lowest simulated-source mean AEE at both $dt=1$ and $dt=4$ horizons under both settings, leading six of eight P and seven of eight C sequence/horizon columns; calibrated DVS-Voltmeter is lower on \texttt{outdoor\_day1} at $dt=1$.
\Cref{fig:flow_qualitative} gives representative outdoor and indoor predictions; see Supplementary Material Sec.~F for the protocol.

\section{Discussion and Conclusion}
\label{sec:conclusion}

We presented \method, a physics-inspired simulator combining fractional-relaxation voltage memory, continuous-time threshold localization, and event-wise reference updates with retained modal state.
Its reusable sensor-side core accepts intensity trajectories from renderers, interpolation modules, and frame sources to determine event timing, polarity, and residual memory.

Under matched inputs, \method\ achieves the lowest mean IEI and Polarity errors, supporting its history-dependent temporal response.
It leads three of six reconstruction metrics on DAVIS240C and HQF, more than any other tested simulator, and yields the lowest simulated-source mean AEE at both flow horizons.
Matching candidate counts aligns the number of evaluated settings, although simulators expose different types and numbers of tunable parameters.
Calibration targets event-stream statistics rather than downstream performance; its effects and gains over simpler dynamics vary by metric.

Sensor-side dynamics cannot recover motion, exposure, or scene information missing from the input trajectory.
The model uses sensor-family global parameters and omits fixed-pattern noise, pixel-threshold mismatch, photon/leakage stochasticity, refractory dead time, and readout arbitration.
Incorporating refractory and readout effects, which affect short-interval firing \cite{joubert2021characterized}, and measuring controlled sensor responses could help relate relaxation modes to pixel circuitry.

Future work includes improved input trajectories, circuit-informed parameter identification, and spatially varying sensor effects.
PECS \cite{pecs2024} and the asynchronous spatiotemporal spike metric \cite{li2023spikemetric} could complement current statistics with joint spatial, temporal, and polarity comparisons of $(x,y,t,p)$ streams with predefined alignment and kernel scales.
Evaluation could address cross-sensor transfer, neuromorphic object detection \cite{li2026neuromorphicdetection}, and event-language understanding such as EventGPT \cite{liu2025eventgpt}.

\clearpage
\onecolumn
\appendix
\makeatletter
\setlength{\@fptop}{0pt}
\makeatother
\renewcommand{\topfraction}{0.95}
\renewcommand{\bottomfraction}{0.80}
\renewcommand{\textfraction}{0.05}
\renewcommand{\floatpagefraction}{0.75}
\setcounter{topnumber}{4}
\setcounter{bottomnumber}{2}
\setcounter{totalnumber}{6}
\setlength{\textfloatsep}{8pt plus 2pt minus 2pt}
\setlength{\floatsep}{8pt plus 2pt minus 2pt}
\setlength{\intextsep}{8pt plus 2pt minus 2pt}
\renewcommand{\theHsection}{appendix.\Alph{section}}
\setcounter{table}{0}
\renewcommand{\thetable}{S\arabic{table}}
\renewcommand{\theHtable}{S\arabic{table}}
\setcounter{figure}{0}
\renewcommand{\thefigure}{S\arabic{figure}}
\renewcommand{\theHfigure}{S\arabic{figure}}
\setcounter{equation}{0}
\renewcommand{\theequation}{S\arabic{equation}}
\renewcommand{\theHequation}{S\arabic{equation}}

\begin{center}
  {\LARGE\bfseries Supplementary Material\par}
\end{center}
\vspace{0.4em}

\begin{center}
  {\bfseries Supplement Scope}
\end{center}
\noindent
In this supplementary material, we provide additional information on input and event-format conventions, dataset roles and splits, hardware and software environment, sensor-family parameter calibration, event-stream comparison details, image reconstruction metrics, qualitative reconstruction examples, and optical-flow estimation details.
The material is organized around model and input conventions and experiment details: model assumptions and input/event-format conventions; dataset roles, splits, and environment; parameter settings and calibration; event-stream comparison details; image reconstruction details; and optical-flow estimation details.
Model assumptions are discussed alongside the input conventions, and evaluation conditions accompany the relevant experiments. The final sections report computational cost and outline further evaluation.
\par\vspace{0.8em}

\section{Model and Input Conventions}
\label{app:input_event_conventions}

\paragraph{Physical motivation.}
Circuit-based DVS models describe finite-bandwidth responses with light-dependent poles, comparator latency, refractory behavior, and readout effects \cite{joubert2021characterized,hu2020v2e}.
FracEvent uses a phenomenological model of distributed relaxation in the pre-threshold signal, represented by a compact positive mixture of exponential modes.
At a fixed operating brightness, the mixture allows fast response and a longer residual tail to coexist; retained states then influence later crossings after reference updates.
The Cole--Cole distribution is a modeling choice; its correspondence to a specific photoreceptor circuit remains to be established.
The first-order, second-order, and uniform-mode controls in Sec.~D compare temporal dynamics under fixed parameter settings.
The observed differences are small and vary across metrics; independent tuning would instead compare the optimized performance of each model family.

\paragraph{Comparison with stochastic voltage models.}
DVS-Voltmeter models voltage drift/diffusion and stochastic first-passage timing \cite{dvsvoltmeter2022}.
For fixed inputs and parameters, FracEvent generates a deterministic, history-dependent voltage trajectory.
This formulation permits repeatable event generation and controlled changes to temporal memory without simulation-seed variability.
Photon and leakage noise are separate stochastic processes that could be incorporated in an extended model.
The evaluated configuration uses zero refractory dead time and omits readout arbitration.
Refractory and readout effects also influence short-interval event activity and are additional mechanisms to consider when modeling repeated firing.

\paragraph{Input Normalization and Trajectory Boundary.}
All methods receive grayscale intensities normalized to a common range before the log transform and use the same intensity floor and numerical epsilon.
We treat gamma correction, demosaicing, exposure reconstruction, color-to-grayscale conversion, high-dynamic-range merging, interpolation, rendering, and target-frame spacing as upstream processing.

\paragraph{Timestamp Convention.}
Frame timestamps, generated event timestamps, MVSEC frame intervals, and reconstruction event windows are all expressed in seconds.

\paragraph{Coordinate and Polarity Convention.}
The raw event tuple uses horizontal coordinate $x$, vertical coordinate $y$, timestamp $t$, and polarity $p$.
Polarity is encoded as $+1$ for ON and $-1$ for OFF.
The OFF threshold parameter is stored as a positive magnitude and is applied through the signed residual condition $d(t)\leq-\thetaoff$.
Downstream event tensors preserve this coordinate and polarity convention.

\section{Dataset Roles, Splits, and Environment}
\label{app:dataset_roles}

\paragraph{Dataset roles.}
The datasets in the main paper serve distinct experimental roles.
Some provide real events for sensor-family calibration or event-stream comparison, some provide high-frame-rate frames for synthetic generation, and some provide real-event targets for downstream transfer.
The protocol separates generation sources, event-statistics calibration, held-out evaluation, and downstream transfer.
When a dataset supports more than one role, these stages use the disjoint windows and signals specified below.

\paragraph{DAVIS240C Role.}
DAVIS240C is used in three distinct ways \cite{brandli2014davis,mueggler2017dataset}.
First, selected DAVIS240C sequences provide APS frames and real events for selecting DAVIS240-series sensor-family parameter settings.
Second, separate held-out DAVIS240C windows are used for event-stream comparison.
Third, full DAVIS240C reconstruction targets are used to test whether a model trained from synthetic GoPro events transfers to real events.
The parameter-selection windows and held-out event-stream windows are disjoint.

\paragraph{DAVIS346 real capture role.}
The DAVIS346 real capture is a small self-collected real-event sample used as additional real-event validation in the main paper.
It provides qualitative event-stream comparison in real-life scenes and supports the DAVIS346-series sensor-regime discussion; optical-flow training and calibration use the MVSEC protocol described below.

\paragraph{GoPro role.}
GoPro provides high-frame-rate intensity input for the reconstruction training source \cite{nah2017gopro}.
Its role is to supply a common brightness trajectory from which \method, v2e \cite{hu2020v2e}, ESIM \cite{rebecq2018esim,video2events2020}, and DVS-Voltmeter \cite{dvsvoltmeter2022} can each generate training streams.
The reconstruction table compares event sources generated from common input scenes.

\paragraph{HQF role.}
HQF is used as an additional real-event reconstruction target \cite{stoffregen2020simtoreal}.
It applies the same reconstruction protocol to different scenes, exposure statistics, motion patterns, and evaluation frames.
Parameter selection uses DAVIS240C event-level signatures, while HQF provides an independent cross-dataset transfer check.

\paragraph{MVSEC role.}
MVSEC supplies DAVIS346-series calibration data and the optical-flow evaluation \cite{mvsec2018}.
The protocol trains each source on \texttt{outdoor\_day2} and evaluates on \texttt{outdoor\_day1} and the indoor flying sequences.
MVSEC also provides DAVIS346-series event-statistics windows for parameter calibration.
The optical-flow experiment tests whether synthetic events support motion learning under a fixed self-supervised protocol, using the DAVIS346-series setting, an event-image representation, an EV-FlowNet-style model, and flow metrics.
\Cref{tab:supp_dataset_roles} summarizes these dataset uses and splits.

\begin{table}[!htbp]
  \centering
  \caption{\textbf{Dataset roles and experimental splits.}}
  \label{tab:supp_dataset_roles}
  \scriptsize
  \setlength{\tabcolsep}{2.6pt}
  \begin{adjustbox}{max width=\linewidth}
  \begin{tabular}{M{0.20\columnwidth}M{0.23\columnwidth}M{0.25\columnwidth}M{0.22\columnwidth}}
    \toprule
    Dataset & Use & Primary output & Split or protocol \\
    \midrule
    DAVIS240C & Calibration, held-out event-stream comparison, real reconstruction target & Sensor-family setting, event-stream table, DAVIS240C reconstruction metrics & Disjoint calibration and held-out event-stream sequences. \\
    DAVIS346 real capture & Additional real-event validation & Real-life event-stream visual comparison & Qualitative validation under the DAVIS346 sensor regime. \\
    GoPro & Common frame source for synthetic training events & Matched simulator comparison & Shared source scenes and target-frame construction. \\
    HQF & Additional real-event reconstruction target & Full-sequence reconstruction metrics & Independent cross-dataset reconstruction target. \\
    MVSEC & DAVIS346-series calibration and optical-flow benchmark & Sensor-family setting, AEE, and outlier metrics & Event-statistics calibration followed by the fixed flow protocol. \\
    \bottomrule
  \end{tabular}
  \end{adjustbox}
\end{table}

\paragraph{Execution environment.}
\Cref{tab:supp_execution_environment} documents the workstation and software environment used for the reported runs.

\begin{table}[!htbp]
  \centering
  \caption{\textbf{Execution environment.}}
  \label{tab:supp_execution_environment}
  \scriptsize
  \setlength{\tabcolsep}{3pt}
  \begin{tabular}{@{}ll@{}}
    \toprule
    Item & Value \\
    \midrule
    CPU & Intel Core i5-14600KF \\
    GPU & NVIDIA RTX 5070 Ti, 16 GB \\
    Python & 3.11.15 \\
    PyTorch & \texttt{2.11.0+cu130} \\
    CUDA & 13.0 \\
    \bottomrule
  \end{tabular}
\end{table}

\section{Parameter Settings and Calibration}
\label{app:calibration_details}

\paragraph{Calibration role.}
Event-camera parameters are sensor- and protocol-dependent.
The paper uses separate parameter settings for the DAVIS240-series and DAVIS346-series sensor families.
These settings are calibrated from real event-stream statistics for the corresponding sensor family.
The calibration objective combines event count, temporal histogram structure, polarity behavior, and time-surface similarity \cite{lagorce2017hots,sironi2018hats}.
The four objective terms jointly constrain event density, timing, polarity, and spatial-temporal recency.
We report public (P) settings for ESIM, v2e, and DVS-Voltmeter alongside calibrated (C) settings selected with equal candidate budgets.
The calibrated comparisons use 240 candidates per family for DAVIS240 and 588 candidates per family for DAVIS346 optical flow.
Calibration data and objective functions are matched across models, while parameter dimensions, search ranges, and capacities remain model-specific.

\paragraph{Calibration objective.}
For each DAVIS240 candidate, the calibration pass compares generated and real event streams on calibration windows with the following local score:
\begin{equation}
  \mathcal{J}_{240} =
  e_{\mathrm{count}}
  +0.5\,e_{\mathrm{temp}}
  +0.35\,e_{\mathrm{pol}}
  +0.25\,(1-\max(c_{\mathrm{surf}},0)).
  \label{eq:supp_calib_objective}
\end{equation}
Here $e_{\mathrm{count}}$ is the Count error, defined as the mean absolute log count ratio; $e_{\mathrm{temp}}$ is a temporal-histogram distance, $e_{\mathrm{pol}}$ measures polarity imbalance, and $c_{\mathrm{surf}}$ is a time-surface correlation.
Count error receives the largest weight to control event-density mismatch.
The other terms constrain timing, polarity, and spatial-temporal recency.
For DAVIS346, let $\bar{\mathcal{J}}_{\mathrm{anchor}}$ and $\bar{\mathcal{J}}_{\mathrm{context}}$ denote the mean local score over 20 activity-stratified \texttt{outdoor\_day2} anchor windows and eight context windows (two each from \texttt{outdoor\_day1} and \texttt{indoor\_flying1/2/3}).
The objective is
\begin{equation}
  \mathcal{J}_{346}=0.8\bar{\mathcal{J}}_{\mathrm{anchor}}
  +0.2\bar{\mathcal{J}}_{\mathrm{context}}
  +\left|\log\frac{N^{A}_{\mathrm{sim}}+1}{N^{A}_{\mathrm{real}}+1}\right|,
  \label{eq:supp_calib_objective_346}
\end{equation}
where $N^{A}$ is the total event count across the anchor windows; the density coefficient is 1.0.
\Cref{tab:supp_calib_terms} summarizes these objective terms and weights.

\begin{table}[!htbp]
  \centering
  \caption{\textbf{Local calibration-score terms.}
  These terms form $\mathcal{J}_{240}$ and the anchor/context scores in $\mathcal{J}_{346}$.}
  \label{tab:supp_calib_terms}
  \scriptsize
  \setlength{\tabcolsep}{3pt}
  \begin{tabular}{M{0.22\columnwidth}M{0.22\columnwidth}M{0.46\columnwidth}}
    \toprule
    Term & Weight & Interpretation \\
    \midrule
    Count error & 1.00 & Penalizes large event-density mismatch across windows. \\
    Temporal histogram L1 & 0.50 & Checks whether events occupy similar temporal portions of the window. \\
    Polarity error & 0.35 & Discourages ON/OFF imbalance that can be hidden by total count. \\
    Time-surface penalty & 0.25 & Balances spatial-temporal recency agreement with the other terms. \\
    \bottomrule
  \end{tabular}
\end{table}

\paragraph{Parameter selection and held-out evaluation.}
Using exhaustive grids, the DAVIS240-series search evaluates 240 candidates and selects $\alpha=0.76$, $M=4$, $\tau_{\mathrm{ref}}=0.0025$ s, $\thetaon=0.5625$, and $\thetaoff=0.4000$.
The DAVIS346-series search exhaustively evaluates 588 candidates under \Cref{eq:supp_calib_objective_346} and selects $\alpha=0.72$, $M=6$, $\tau_{\mathrm{ref}}=0.0025$ s, $\thetaon=0.4500$, and $\thetaoff=0.5000$.
The DAVIS240C experiments use the DAVIS240-series setting, while the MVSEC experiments use the DAVIS346-series setting.
\Cref{tab:supp_calibration} summarizes the selected sensor-family settings; downstream results are reported in the main paper.

\paragraph{Calibration and evaluation splits.}
The DAVIS240C search uses four calibration sequences: \texttt{boxes\_6dof}, \texttt{dynamic\_rotation}, \texttt{poster\_translation}, and \texttt{slider\_depth}.
Each candidate is evaluated on the first 40 APS frames of those calibration sequences.
Held-out event-stream evaluation uses all APS frames from the disjoint sequences \texttt{boxes\_rotation}, \texttt{dynamic\_translation}, \texttt{poster\_rotation}, and \texttt{slider\_far}, with 64 deterministic 50 ms windows spread over each sequence.
The DAVIS346-series search uses the 20 anchor and eight context windows defined above to match event density and local event statistics.
The downstream reconstruction models are then trained on GoPro-generated events and evaluated on the full DAVIS240C and HQF targets.
The full downstream optical-flow table is trained and evaluated with the fixed MVSEC protocol described in \Cref{app:mvsec_details}.
Parameter selection uses the event-level calibration objective before downstream data generation and training.

\paragraph{Sensor-family scope.}
The reported parameter sets correspond to the DAVIS240- and DAVIS346-series sensors and capture protocols.
Other cameras, bias settings, lenses, exposure regimes, or upstream trajectory sources may require recalibration.

\paragraph{DAVIS240 event-stream protocol.}
The simulator comparison selects FracEvent, ESIM, v2e, and DVS-Voltmeter with the shared event-only objective in \Cref{eq:supp_calib_objective}, using the fixed 240-candidate grids in \Cref{tab:supp_downstream_search_240}.
Stochastic DVS-Voltmeter evaluations use one fixed simulation seed for reproducibility.
For held-out evaluation, every selected simulator is advanced through all APS frames of \texttt{boxes\_rotation}, \texttt{dynamic\_translation}, \texttt{poster\_rotation}, and \texttt{slider\_far}.
We score 64 deterministic 50 ms windows spread over the full duration of each sequence (256 windows per setting).
Besides Count, IEI, Polarity, and Time Surface, the evaluation reports normalized per-pixel Spatial error, polarity-separated normalized Voxel error with 10 temporal bins and $6\!\times\!6$-pixel spatial cells, and Repeated IEI error for the fraction of same-pixel IEIs no longer than 10 ms.
Calibration and held-out evaluation use disjoint windows.
All simulator families share the calibration data, objective, candidate count, and held-out windows.
\Cref{tab:supp_downstream_search_240} lists their native parameterizations.

\paragraph{Ablation protocol.}
All temporal and component variants use FracEvent's calibrated DAVIS240 parameter setting: $\alpha=0.76$, $M=4$, $\tau_{\rm ref}=0.0025$ s, $\theta_{\rm ON}=0.5625$, and $\theta_{\rm OFF}=0.4000$.
Each variant changes one named mechanism.
The first-order variant sets $M=1$ and $w_1=1$ while retaining the selected $\tau_{\rm ref}$ and thresholds; $\alpha$ is not used.
The second-order variant uses the real-pole cascade $H(s)=1/[(1+\tau_1s)(1+\tau_2s)]$, with $\tau_1=\tau_0/\sqrt{5}$ and $\tau_2=\tau_0\sqrt{5}$.
Its mode representation uses the exact partial-fraction weights $w_1=\tau_1/(\tau_1-\tau_2)$ and $w_2=-\tau_2/(\tau_1-\tau_2)$, providing a concrete two-timescale second-order baseline.
Uniform-4 retains the four log-time nodes over $[-3,3]$ and replaces the fractional quadrature weights by equal weights of $1/4$.
The interval-reset ablation resets all modal states to their current references after each event-bearing APS-frame interval.
The shared-threshold ablation uses the arithmetic mean of the calibrated thresholds, $\theta_{\rm ON}=\theta_{\rm OFF}=0.48125$.
All variants use the same held-out sequences, windows, inputs, and event metrics described above.

\paragraph{Downstream calibration and training protocol.}
Within each downstream task, all simulator families share the calibration data, objective, candidate count, and downstream training protocol, while retaining their native searched variables.
Each selected setting generates the training source used by the fixed downstream protocol.
For DAVIS240 reconstruction, every family receives 240 event-only candidate evaluations under \Cref{eq:supp_calib_objective}.
FracEvent searches its fractional parameters and thresholds; DVS-Voltmeter searches its model-specific parameter scales; ESIM and v2e search only ON/OFF thresholds, with refractory fixed to zero and v2e cutoff, threshold noise, leak, shot noise, and photoreceptor noise disabled.
Tuning the v2e intensity-dependent cutoff would extend this search to bandwidth parameters.
For DAVIS346 optical flow, every family receives 588 event-only evaluations under \Cref{eq:supp_calib_objective_346}.
ESIM and v2e use one fixed $21\!\times\!28$ threshold grid over $[0.12,1.20]$ with the same optional effects disabled, selecting $(1.20,1.20)$ and $(0.444,0.840)$; DVS-Voltmeter uses one fixed 588-point grid over its model-specific parameters; FracEvent uses its 588-candidate setting in \Cref{tab:supp_calibration}.

\begin{table}[!htbp]
  \centering
  \caption{\textbf{DAVIS240-series calibration grids.}
  Every simulator evaluates one fixed 240-candidate grid on the same calibration sequences and event-only objective.}
  \label{tab:supp_downstream_search_240}
  \scriptsize
  \setlength{\tabcolsep}{2.2pt}
  \begin{adjustbox}{max width=\linewidth}
  \begin{tabular}{llll}
    \toprule
    Family & Searched grid & Selected setting & Candidates \\
    \midrule
    \method & \shortstack[l]{$\alpha\!:\{.76,.78,.80,.82,.84\}$; $M=4$\\$\tau\!:\{.0015,.0020,.0025,.0030\}$\\ON: $\{.4500,.4875,.5250,.5625\}$; OFF: $\{.4000,.4500,.5000\}$} & $\alpha=.76$, $M=4$, $\tau=.0025$, ON/OFF $.5625/.4000$ & 240 \\
    DVS-Voltmeter & $k_1$: 5 $[.65,1.35]$; $k_2$: 4 $[.65,1.50]$; $k_3$: 3 $[.25,2]$; noise: 4 $[0,2]$ & scales $1.00/.65/.25/.00$ & 240 \\
    v2e & ON: 15 values $[.12,1.20]$; OFF: 16 values $[.12,1.20]$ & ON/OFF $.660/.480$; optional effects off & 240 \\
    ESIM & ON: 15 values $[.12,1.20]$; OFF: 16 values $[.12,1.20]$ & ON/OFF $1.200/1.128$; refractory 0 & 240 \\
    \bottomrule
  \end{tabular}
  \end{adjustbox}
\end{table}

\begin{table}[!htbp]
  \centering
  \caption{\textbf{DAVIS346-series calibration grids.}
  Every simulator evaluates one fixed 588-candidate grid on the same calibration windows and event-only objective.}
  \label{tab:supp_downstream_search_588}
  \scriptsize
  \setlength{\tabcolsep}{2.2pt}
  \begin{adjustbox}{max width=\linewidth}
  \begin{tabular}{llll}
    \toprule
    Family & Searched grid & Selected setting & Candidates \\
    \midrule
    \method & $\alpha$: 3 $[.72,.80]$; $M$: 2 $\{4,6\}$; $\tau$: 2 $[.0025,.004]$; ON: 7 $[.30,.60]$; OFF: 7 $[.25,.55]$ & $\alpha=.72$, $M=6$, $\tau=.0025$, ON/OFF $.450/.500$ & 588 \\
    DVS-Voltmeter & $k_1$: 7 $[.65,1.35]$; $k_2$: 7 $[.65,1.50]$; $k_3$: 3 $[.50,2]$; noise: 4 $[0,2]$ & scales $.80/.65/2.00/1.00$ & 588 \\
    v2e & ON: 21 values $[.12,1.20]$; OFF: 28 values $[.12,1.20]$ & ON/OFF $.444/.840$; optional effects off & 588 \\
    ESIM & ON: 21 values $[.12,1.20]$; OFF: 28 values $[.12,1.20]$ & ON/OFF $1.200/1.200$; refractory 0 & 588 \\
    \bottomrule
  \end{tabular}
  \end{adjustbox}
\end{table}

\begin{table}[!htbp]
  \centering
  \caption{\textbf{Sensor-family parameter settings.}
  Calibration source, search scope, and selected parameters for the two sensor families.}
  \label{tab:supp_calibration}
  \scriptsize
  \setlength{\tabcolsep}{3pt}
  \begin{tabular}{M{0.30\columnwidth}M{0.29\columnwidth}M{0.29\columnwidth}}
    \toprule
    Item & DAVIS240-series & DAVIS346-series \\
    \midrule
    Calibration source & DAVIS240C windows & MVSEC windows \\
    Window plan & Four sequences, first 40 APS frames & 20 \texttt{outdoor\_day2} anchors + 8 context windows (two from each of four other sequences) \\
    Candidates evaluated & 240 & 588 \\
    $\alpha$ and modes $M$ & 0.76, 4 & 0.72, 6 \\
    $\tau_{\mathrm{ref}}$ & 0.0025 s & 0.0025 s \\
    $\thetaon$ and $\thetaoff$ & 0.5625, 0.4000 & 0.4500, 0.5000 \\
    \bottomrule
  \end{tabular}
\end{table}

\paragraph{Fixed implementation settings.}
In addition to the five searched sensor-family parameters in \Cref{tab:supp_calibration}, \method\ uses fixed implementation settings shared by both reported profiles.
\Cref{tab:supp_fixed_settings} lists these implementation constants and input conventions.

\begin{table}[!htbp]
  \centering
  \caption{\textbf{Fixed implementation settings for \method.}
  The listed values are shared across the reported DAVIS240-series and DAVIS346-series profiles.}
  \label{tab:supp_fixed_settings}
  \scriptsize
  \setlength{\tabcolsep}{3pt}
  \begin{tabular}{M{0.31\columnwidth}M{0.25\columnwidth}M{0.34\columnwidth}}
    \toprule
    Setting & Value & Role \\
    \midrule
    Relaxation log range $R$ & 3.0 & Samples log-time coordinates $\xi_m\in[-R,R]$ for $\tau_m=\tau_0 e^{\xi_m}$. \\
    $\tau_{\min},\tau_{\max}$ & $5{\times}10^{-5}$ s, $3.0{\times}10^{-2}$ s & Bounds the brightness-dependent base time constant. \\
    $I_{\mathrm{ref}},\beta$ & 0.5, 1.0 & Sets the brightness dependence of the base time constant. \\
    $i_{\mathrm{dark}},\epsilon$ & 0.001, $10^{-6}$ & Stabilizes log-intensity normalization near zero intensity. \\
    Bisection steps & 12 & Root-localization iterations for counted threshold levels. \\
    Solver cap, minimum split step & 64 events per interval, $10^{-6}$ s & Controls recursive splitting in dense intervals. \\
    Input convention & Normalized grayscale, seconds, piecewise-linear $L(t)$ & Shared timestamp, intensity, and interpolation convention. \\
    \bottomrule
  \end{tabular}
\end{table}

\paragraph{Threshold-solver condition.}
The event-generation solver counts threshold levels spanned by the endpoint voltage and then localizes each counted level by bisection.
This root-localization step is exact for a pixel on a processed subinterval where the weighted voltage trajectory is monotone, because each counted threshold level has a unique crossing.
The carried fractional mode state can make a long interval's weighted voltage non-monotone.
Dense intervals are therefore recursively split before bisection, applying endpoint counting on shorter subintervals without a uniform substep grid.

\section{Event-Stream Comparison Details}
\label{app:event_stream_details}

\begin{figure}[!htbp]
  \centering
  \includegraphics[width=\linewidth,height=0.78\textheight,keepaspectratio]{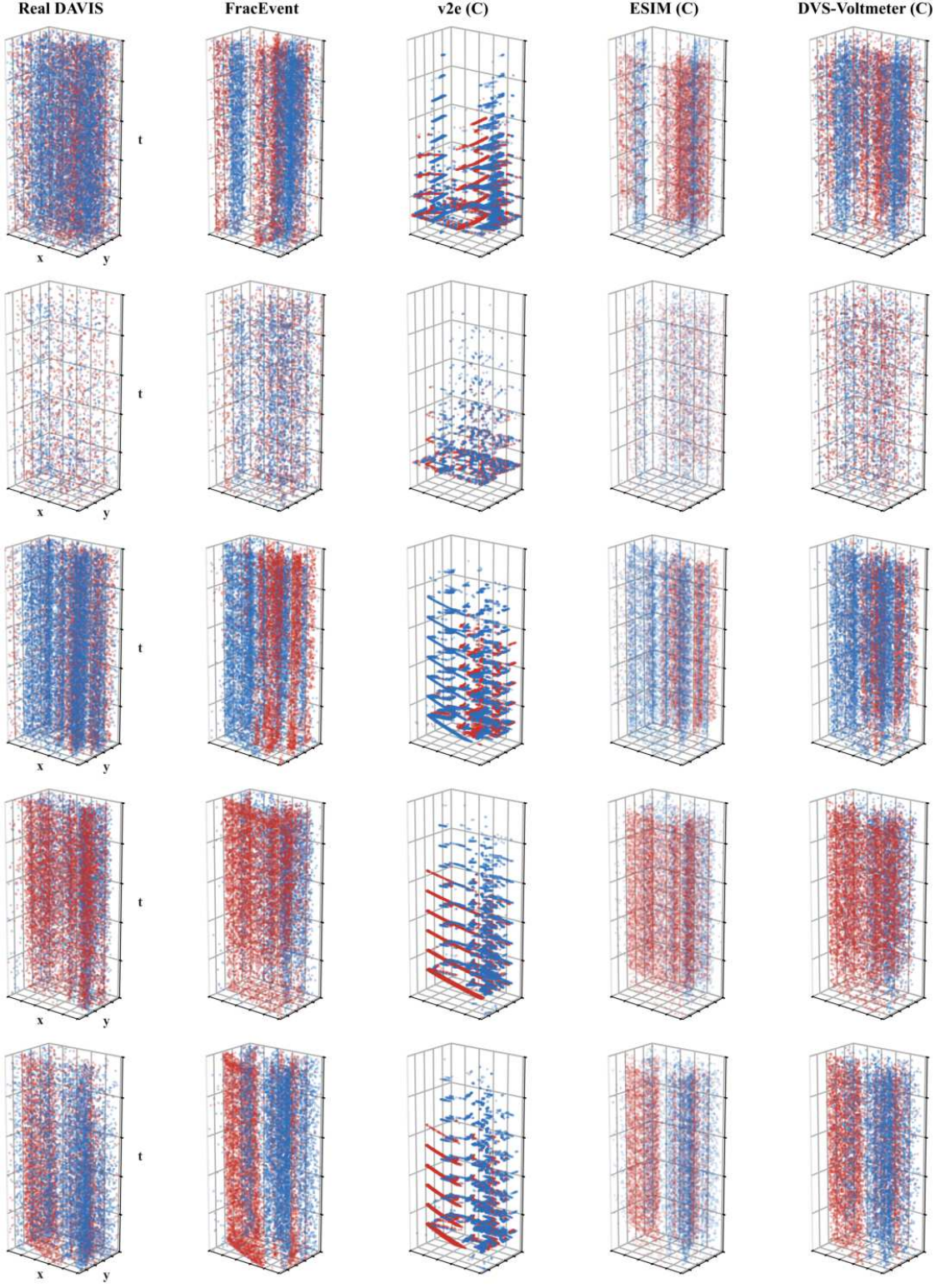}
  \caption{\textbf{Additional qualitative event-stream comparisons.}
  Each row shows a spatiotemporal event plot for one 50 ms DAVIS240C window, with columns comparing real events and generated events from the same interval. C denotes the 240-candidate calibrated setting; red and blue denote ON and OFF events.}
  \label{fig:supp_event_stream_extra}
\end{figure}

\paragraph{Matched-window protocol.}
The DAVIS240C event-stream comparison uses the four held-out sequences listed in \Cref{app:calibration_details}.
For each sequence, 64 deterministic frame-aligned 50 ms windows are spread over the full APS duration, giving 256 matched windows in total.
The windows are fixed before scoring and are shared by all simulator baselines and controlled dynamics.
All event sources are scored against the same real DAVIS240C windows with identical start and end timestamps, spatial resolution, timestamp units, polarity convention, and metric implementation.
For the controlled dynamics comparison, every variant inherits the same calibrated FracEvent parameter setting, input frames, threshold/reference update, root-localization implementation, matched windows, and metrics; only the named temporal model family changes.
\Cref{fig:supp_event_stream_extra} shows additional qualitative event-stream comparisons under this fixed-window protocol.

\paragraph{Metric details.}
For a simulated event window $S$ and the matched real event window $R$, the count ratio is:
\begin{equation}
  r_{\mathrm{count}}(S,R)=\frac{|S|+1}{|R|+1}.
  \label{eq:supp_event_count_ratio}
\end{equation}
The additive one keeps empty or near-empty windows numerically defined.
The reported Count error is the absolute logarithm
\begin{equation}
  e_{\mathrm{count}}(S,R)=\left|\log r_{\mathrm{count}}(S,R)\right|,
  \label{eq:supp_event_count_log}
\end{equation}
which treats reciprocal over- and under-production symmetrically.

Let $\mathcal{I}(E)$ denote the positive same-pixel consecutive intervals obtained after sorting an event stream $E$ by pixel and timestamp.
The IEI distance clips intervals below $10^{-6}$ s before the logarithm and applies the 1-Wasserstein distance \cite{villani2009optimaltransport}:
\begin{equation}
\begin{aligned}
  \tilde{\mathcal{I}}(E) &=
  \log_{10}\!\left(\max(\mathcal{I}(E),10^{-6})\right),\\
  d_{\mathrm{IEI}}(S,R) &=
  W_1\!\left(\tilde{\mathcal{I}}(S),\tilde{\mathcal{I}}(R)\right).
\end{aligned}
  \label{eq:supp_event_iei}
\end{equation}
Windows without valid same-pixel intervals in either stream are excluded from both interval-based aggregates.

Polarity is the absolute difference between simulated and real ON-event fractions:
\begin{equation}
  \begin{aligned}
  \pi_+(E) &=
  \begin{cases}
  |E|^{-1}\sum_{e\in E}\mathbf{1}[p_e=+1], & |E|>0,\\
  0.5, & |E|=0,
  \end{cases}\\
  d_{\mathrm{pol}}(S,R) &= \left|\pi_+(S)-\pi_+(R)\right|.
  \end{aligned}
  \label{eq:supp_event_polarity}
\end{equation}

For Time Surface \cite{lagorce2017hots,sironi2018hats}, let $t_p^E(x,y)$ be the latest timestamp of polarity $p$ at pixel $(x,y)$ in stream $E$, and set the corresponding surface entry to zero when no such event exists.
Using $\tau=0.05$ s and the common window end time $t_1$, we compute:
\begin{equation}
\begin{aligned}
  T_p^E(x,y) &=
  \exp\!\left(-\frac{t_1-t_p^E(x,y)}{\tau}\right),\\
  d_{\mathrm{surf}}(S,R) &=
  1-\max\!\left(
  \frac{1}{|\mathcal{P}_v|}
  \sum_{p\in\mathcal{P}_v}
  \rho\!\left(T_p^S,T_p^R\right),0\right),
\end{aligned}
  \label{eq:supp_event_time_surface}
\end{equation}
where $\mathcal{P}_v$ contains polarities with finite correlations.
For Spatial, let $h_E(x,y)=n_E(x,y)/\max(|E|,1)$ be the per-pixel event-count map normalized by the total number of events in the window.
We report
\begin{equation}
  d_{\mathrm{spatial}}(S,R)=\left\|h_S-h_R\right\|_1.
  \label{eq:supp_event_spatial}
\end{equation}
Spatial therefore measures where activity occurs after marginalizing over time and polarity.
For the complementary Voxel measure, events are separated by polarity and accumulated into 10 temporal bins and $6\!\times\!6$-pixel spatial cells.
Each grid $V(E)$ is normalized by its event count, and we report
\begin{equation}
  d_{\mathrm{voxel}}(S,R)=\left\|V(S)-V(R)\right\|_1.
  \label{eq:supp_event_voxel}
\end{equation}
Finally, for nonempty $\mathcal{I}(E)$, let $q(E)=|\{\Delta t\in\mathcal{I}(E):\Delta t\leq10\,\mathrm{ms}\}|/|\mathcal{I}(E)|$.
We report $|q(S)-q(R)|$ as the \emph{Repeated IEI} error.
The metric set covers density, repeated-trigger timing, polarity balance, and recent-event spatial structure under one fixed window protocol.

\paragraph{Full-sequence results.}
All entries in \Cref{tab:supp_dynamics_results,tab:supp_component_ablation,tab:supp_public_matched_results} are held-out means.
IEI and Repeated IEI averages use 255 valid paired windows because the final \texttt{slider\_far} real window has no valid same-pixel interval for any setting; the other metrics use all 256 windows.
The descriptive intervals below resample the four sequences as clusters 10,000 times.
With four sequence-level units, these intervals describe sequence-level variation.
They quantify variation in event-stream statistics across sequences; variability across network-training seeds would require repeated downstream training runs.

\begin{table}[!htbp]
  \centering
  \caption{\textbf{Temporal-dynamics comparison.}
  All variants use the same FracEvent parameter setting and differ only in their temporal dynamics. Mean held-out errors; lower is better.}
  \label{tab:supp_dynamics_results}
  \scriptsize
  \setlength{\tabcolsep}{1.6pt}
  \begin{adjustbox}{max width=\linewidth}
  \begin{tabular}{lrrrrrrr}
    \toprule
    Dynamics & Count & IEI & Polarity & Time surf. & Spatial & Voxel & Repeated IEI \\
    \midrule
    Fractional & 0.6535 & \best{0.0978} & \best{0.0303} & 0.6284 & \best{0.9693} & \best{1.1201} & 0.0776 \\
    First-order & \best{0.6256} & 0.1010 & 0.0475 & \best{0.6124} & 0.9726 & 1.1243 & 0.1050 \\
    Second-order & 0.7815 & 0.1174 & 0.0624 & 0.7212 & 1.0920 & 1.2455 & 0.1520 \\
    Uniform-4 & 0.6877 & 0.1134 & 0.0445 & 0.6409 & 0.9909 & 1.1232 & \best{0.0721} \\
    \bottomrule
  \end{tabular}
  \end{adjustbox}
\end{table}

For paired mean IEI (alternative minus fractional), first-order is $+0.0031$ [$-0.0220,0.0361$], second-order is $+0.0196$ [$-0.0098,0.0463$], and Uniform-4 is $+0.0156$ [$-0.0006,0.0253$].
Thus fractional has the lowest mean IEI, but none of these descriptive sequence-cluster intervals excludes zero.
Other metrics expose trade-offs: first-order lowers Time Surface error by $0.0160$ [$0.0120,0.0199$] but raises Repeated IEI by $0.0273$ [$0.0045,0.0503$]; Uniform-4 has the lowest Repeated IEI point estimate, whereas fractional has the lowest Polarity, Spatial, and Voxel point estimates.

\begin{table}[!htbp]
  \centering
  \caption{\textbf{Component ablation.}
  All variants use the calibrated FracEvent parameter set. ``w/o multi-mode memory'' retains one first-order state; ``w/o retained memory state'' resets the modes after event-bearing APS intervals; and ``w/o separate thresholds'' replaces the ON/OFF thresholds by their mean. Mean held-out errors; lower is better.}
  \label{tab:supp_component_ablation}
  \scriptsize
  \setlength{\tabcolsep}{1.6pt}
  \begin{adjustbox}{max width=\linewidth}
  \begin{tabular}{lrrrrrrr}
    \toprule
    Variant & Count & IEI & Polarity & Time surf. & Spatial & Voxel & Repeated IEI \\
    \midrule
    Full FracEvent & 0.6535 & \best{0.0978} & \best{0.0303} & 0.6284 & \best{0.9693} & \best{1.1201} & 0.0776 \\
    w/o multi-mode memory & \best{0.6256} & 0.1010 & 0.0475 & \best{0.6124} & 0.9726 & 1.1243 & 0.1050 \\
    w/o retained memory state & 0.6566 & 0.1047 & 0.0492 & 0.6298 & 0.9947 & 1.1548 & 0.1016 \\
    w/o separate thresholds & 0.6598 & 0.1070 & 0.0987 & 0.6285 & 0.9752 & 1.1398 & \best{0.0762} \\
    \bottomrule
  \end{tabular}
  \end{adjustbox}
\end{table}

Relative to Full FracEvent, the variant w/o retained memory state has higher Polarity error ($0.0492$ versus $0.0303$), and raises Spatial by $0.0254$ [$0.0219,0.0296$], Voxel by $0.0346$ [$0.0263,0.0415$], and Repeated IEI by $0.0239$ [$0.0029,0.0452$].
The variant w/o separate thresholds has higher Polarity error ($0.0987$ versus $0.0303$), and raises IEI by $0.0092$ [$0.0017,0.0146$] and Voxel by $0.0197$ [$0.0125,0.0305$].
The ablations isolate distinct event-statistic effects from cross-interval modal carry-over and calibrated ON/OFF asymmetry.

\begin{table}[!htbp]
  \centering
  \caption{\textbf{Effect of calibration on event-stream metrics.}
  Held-out means; C/P denote 240-candidate calibrated/public settings.}
  \label{tab:supp_public_matched_results}
  \scriptsize
  \setlength{\tabcolsep}{1.4pt}
  \begin{adjustbox}{max width=\linewidth}
  \begin{tabular}{lrrrrrrr}
    \toprule
    Setting & Count & IEI & Polarity & Time surf. & Spatial & Voxel & Repeated IEI \\
    \midrule
    \method\ (C=P) & 0.6535 & \best{0.0978} & \best{0.0303} & 0.6284 & 0.9693 & 1.1201 & \best{0.0776} \\
    ESIM (C) & 0.6649 & 0.2176 & 0.0641 & 0.6462 & 1.2313 & 1.1800 & 0.3289 \\
    ESIM (P) & 1.6635 & 0.6151 & 0.0790 & 0.5983 & 1.0669 & 1.0361 & 0.3082 \\
    v2e (C) & \best{0.5341} & 0.2741 & 0.0471 & 0.5970 & 0.9226 & 1.3361 & 0.3174 \\
    v2e (P) & 1.0793 & 0.4460 & 0.0816 & 0.7425 & \best{0.8171} & 1.1610 & 0.2508 \\
    DVS-Voltmeter (C) & 0.6020 & 0.1915 & 0.0811 & \best{0.5609} & 0.8310 & \best{1.0170} & 0.1891 \\
    DVS-Voltmeter (P) & 0.6912 & 0.2281 & 0.1458 & 0.5919 & 0.8890 & 1.0822 & 0.3346 \\
    \bottomrule
  \end{tabular}
  \end{adjustbox}
\end{table}

Calibration lowers IEI for all three baselines.
Voxel error rises for ESIM and v2e and falls for DVS-Voltmeter.
Calibrated-minus-public IEI differences are $-0.3975$ [$-0.4781,-0.3175$] for ESIM, $-0.1718$ [$-0.2166,-0.1356$] for v2e, and $-0.0367$ [$-0.0475,-0.0214$] for DVS-Voltmeter.
The corresponding Voxel differences are $+0.1439$ [$0.0942,0.1935$], $+0.1751$ [$0.1430,0.2072$], and $-0.0652$ [$-0.0861,-0.0503$].
These differences show objective-dependent calibration effects across simulator-native parameter spaces.

\paragraph{Repeated-trigger statistics in real events.}
Real event streams contain frequent same-pixel repeats, with capture-dependent variation.
These measurements characterize repeated-trigger and short-interval burst structure in real event streams, providing context for the retained-memory and continuous-time crossing behavior evaluated by the event-stream and temporal-dynamics tables.
The analysis samples 16 deterministic, uniformly spaced 50 ms windows per capture and reports median [IQR] percentages across windows, where IQR is the 25th--75th percentile range.
Repeated-pixel ratio is the fraction of active pixels that fire at least twice in the window.
Burst-IEI ratio is the fraction of same-pixel consecutive event intervals below 10 ms.
Multi-event contribution is the fraction of all events emitted by pixels that fire at least twice in the same window.
\Cref{tab:supp_repeated_trigger_analysis} reports sampled real-event evidence across DAVIS240C and MVSEC captures.

\begin{table}[!htbp]
  \centering
  \caption{\textbf{Per-capture repeated-trigger analysis.}
  Entries are median [IQR] percentages over sampled 50 ms real-event windows; IQR is the 25th--75th percentile range.}
  \label{tab:supp_repeated_trigger_analysis}
  \scriptsize
  \setlength{\tabcolsep}{2pt}
  \renewcommand{\arraystretch}{0.95}
  \begin{adjustbox}{max width=\linewidth}
  \begin{tabular}{M{0.34\columnwidth}rrrr}
    \toprule
    Capture & Win. & Rep. px & IEI $<10$ms & Rep. events \\
    \midrule
    DAVIS240C \texttt{boxes\_rotation} & 16 & 89.6 [68.8, 95.1] & 73.1 [62.8, 79.8] & 97.7 [87.2, 99.1] \\
    DAVIS240C \texttt{dynamic\_translation} & 16 & 62.2 [54.9, 65.5] & 71.4 [69.9, 73.7] & 83.9 [78.4, 86.3] \\
    DAVIS240C \texttt{poster\_rotation} & 16 & 83.2 [65.5, 95.4] & 72.8 [63.3, 79.8] & 94.6 [85.9, 99.1] \\
    DAVIS240C \texttt{slider\_far} & 16 & 41.8 [37.1, 47.0] & 51.9 [46.5, 57.3] & 64.3 [59.5, 69.9] \\
    MVSEC \texttt{indoor\_flying1} & 16 & 10.3 [8.6, 15.5] & 0.4 [0.2, 1.3] & 19.2 [16.6, 27.5] \\
    MVSEC \texttt{indoor\_flying2} & 16 & 13.7 [10.1, 17.0] & 0.4 [0.2, 0.9] & 24.7 [19.1, 29.8] \\
    MVSEC \texttt{indoor\_flying3} & 16 & 12.8 [10.8, 14.6] & 0.4 [0.2, 0.5] & 23.3 [20.2, 26.2] \\
    MVSEC \texttt{outdoor\_day1} & 16 & 39.4 [32.0, 46.6] & 61.1 [57.2, 63.2] & 64.9 [55.1, 70.9] \\
    MVSEC \texttt{outdoor\_day2} & 16 & 50.0 [36.6, 62.4] & 59.9 [57.2, 67.5] & 74.0 [59.0, 85.0] \\
    \bottomrule
  \end{tabular}
  \end{adjustbox}
\end{table}

\section{Image Reconstruction Details}
\label{app:reconstruction_details}

\paragraph{Training protocol.}
The reconstruction experiment trains the same E2VID-style recurrent model \cite{rebecq2019e2vid} from synthetic GoPro events and evaluates each trained model on real event streams.
All simulated sources share the same GoPro input, $240\times180$ training resolution, target-frame construction, loss, and optimizer settings.
The protocol fixes training at 20 epochs; source-dependent valid-window counts produce slightly different optimizer-update totals.
Evaluation uses the full DAVIS240C and HQF sequences.
Each reconstruction setting is trained once; the tables report transfer performance from that run.

\paragraph{Evaluation scope.}
The reconstruction scores measure the usefulness of simulated training events for transfer to real events, with recorded APS frames as intensity targets.
They reflect agreement with these targets, including their brightness.
Evaluating latent photocurrent fidelity or low-light enhancement requires separate reference measurements and protocols.

\paragraph{Evaluation granularity.}
The evaluation averages over the full target sequences: 22,680 DAVIS240C frames and 15,498 HQF frames in the matched GoPro comparison.
Full-sequence averaging matters because qualitative reconstructions can vary sharply with scene texture, illumination, and motion.

\paragraph{Reconstruction metrics.}
Let $I_t,\hat I_t\in[0,1]^N$ denote the target and reconstructed grayscale frames over $N$ valid pixels.
The main reconstruction table reports full-sequence averages of MSE, SSIM, and LPIPS, following event-reconstruction reporting practice \cite{evreal2023}.
The pixelwise error metric is:
\begin{equation}
  \mathrm{MSE}_t =
  \frac{1}{N}\sum_{i=1}^{N}(\hat I_{t,i}-I_{t,i})^2.
  \label{eq:supp_recon_mse}
\end{equation}
SSIM measures luminance, contrast, and structural agreement \cite{wang2004ssim}:
\begin{equation}
  \mathrm{SSIM}(\hat I,I)=
  \frac{(2\mu_{\hat I}\mu_I+C_1)(2\sigma_{\hat I I}+C_2)}
       {(\mu_{\hat I}^2+\mu_I^2+C_1)(\sigma_{\hat I}^2+\sigma_I^2+C_2)}.
  \label{eq:supp_recon_ssim}
\end{equation}
LPIPS compares learned deep features with calibrated channel weights \cite{zhang2018lpips}:
\begin{equation}
  \mathrm{LPIPS}(\hat I,I)=
  \sum_l\frac{1}{H_lW_l}\sum_{h,w}
  \left\|w_l\odot(\phi_l(\hat I)_{hw}-\phi_l(I)_{hw})\right\|_2^2.
  \label{eq:supp_recon_lpips}
\end{equation}
Lower MSE and LPIPS indicate better agreement, while higher SSIM indicates stronger structural similarity.

\paragraph{Event windows and target alignment.}
The reconstruction learner receives event tensors built from raw event streams and learns to predict intensity targets at the paired frame timestamps.
This pairing is part of the protocol: changing the number of frames between targets changes the temporal integration problem even if the same sensor-family setting is used.
For this reason, the main simulator comparison fixes the source split, target-frame construction, spatial resolution, optimizer settings, 20-epoch schedule, and evaluation code.

\paragraph{Primary reconstruction target.}
The image-reconstruction comparison uses the DAVIS240-series setting because its primary real-event target is DAVIS240C.
HQF tests cross-dataset transfer under the same GoPro training protocol with different scenes, exposure statistics, motion patterns, and evaluation frames.
The main paper reports the public and calibrated results, with additional protocol details and measurements given below.

\begin{figure}[!htbp]
  \centering
  \includegraphics[width=0.92\textwidth]{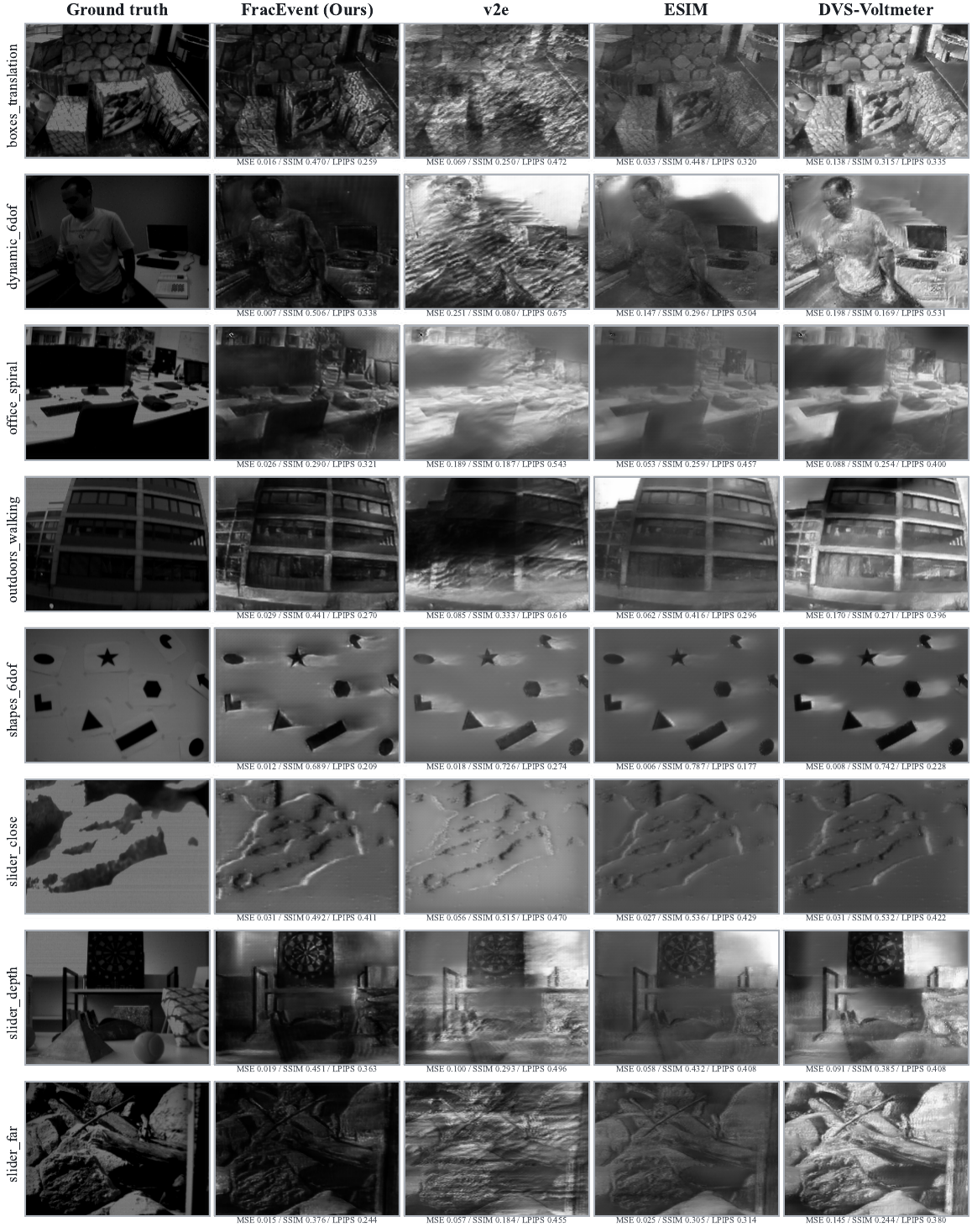}
  \caption{\textbf{Additional DAVIS240C reconstruction comparisons.}
  Additional scenes evaluated with the same GoPro-trained models and raw $[0,1]$ display convention as the main figure.}
  \label{fig:supp_recon_davis}
\end{figure}

\begin{figure}[!htbp]
  \centering
  \includegraphics[width=0.85\textwidth]{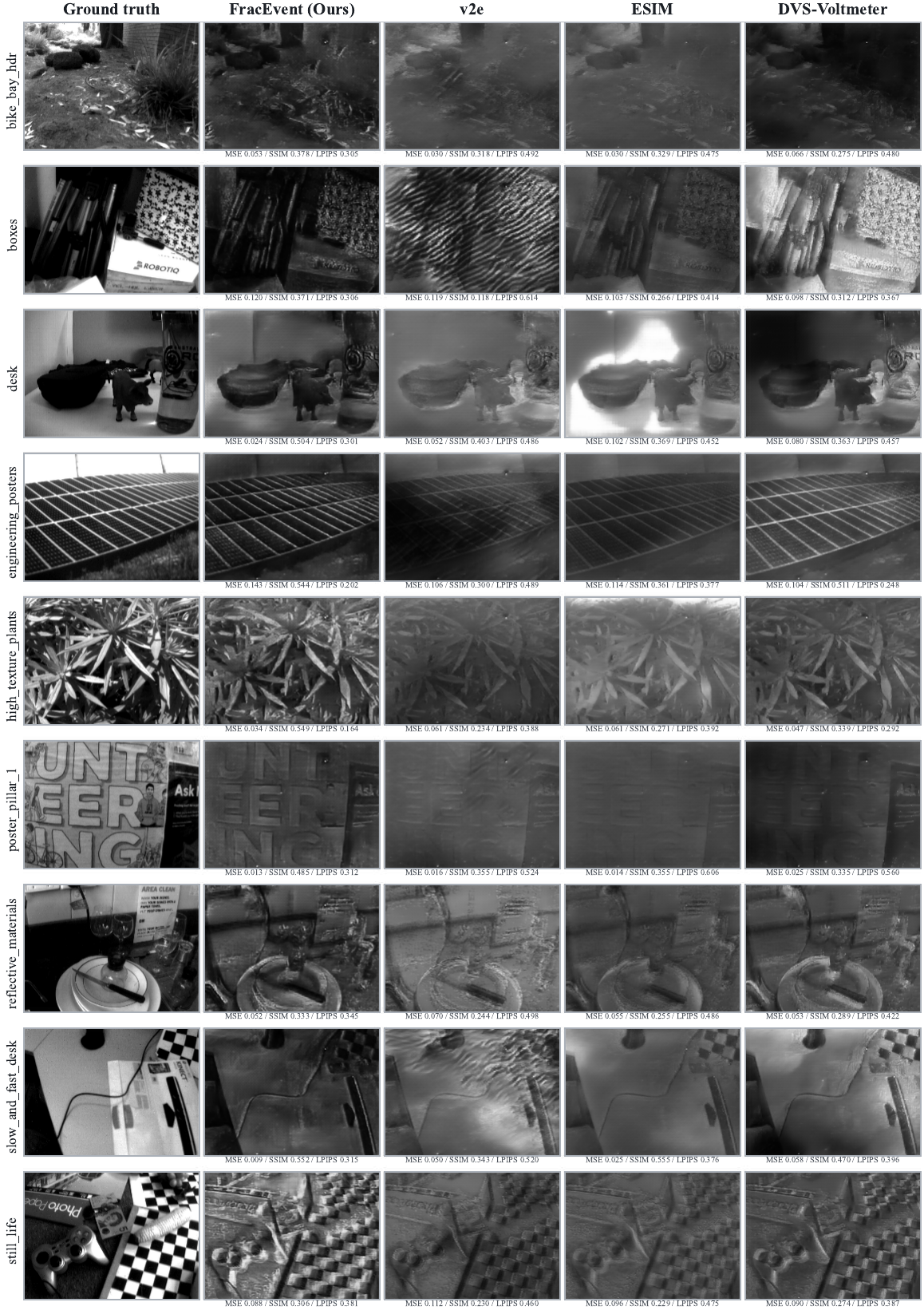}
  \caption{\textbf{Additional HQF reconstruction comparisons.}
  HQF scenes under the same reconstruction protocol and display convention as the DAVIS240C comparison.}
  \label{fig:supp_recon_hqf}
\end{figure}

\paragraph{Qualitative reconstruction examples.}
\Cref{fig:supp_recon_davis,fig:supp_recon_hqf} extend the public-setting visual comparison to additional DAVIS240C and HQF scenes under the same raw-display convention.

\paragraph{Brightness statistics.}
The main-paper reconstruction panels show raw $[0,1]$ outputs, allowing direct comparison of output brightness.
\Cref{tab:supp_brightness} reports brightness statistics for the target and reconstructed outputs.
These raw-output statistics describe brightness agreement under the stated display convention.

\begin{table}[!htbp]
  \centering
  \caption{\textbf{Brightness statistics for the main-paper reconstruction examples.}
  Values are measured from the raw $[0,1]$ outputs shown in the main-paper reconstruction figure.}
  \label{tab:supp_brightness}
  \scriptsize
  \setlength{\tabcolsep}{2.4pt}
  \begin{tabular}{lrrrr}
    \toprule
    Source & Mean & P95 & White & Black \\
    \midrule
    Target & 0.1038 & 0.3020 & 0.0000 & 0.3340 \\
    \method & 0.1005 & 0.2235 & 0.0000 & 0.2196 \\
    v2e & 0.4424 & 0.9255 & 0.0381 & 0.0112 \\
    ESIM & 0.3532 & 0.9647 & 0.0583 & 0.0001 \\
    DVS-Voltmeter & 0.5359 & 0.8078 & 0.0014 & 0.0000 \\
    \bottomrule
  \end{tabular}
\end{table}

\section{Optical-Flow Estimation Details}
\label{app:mvsec_details}

\paragraph{Optical-flow transfer role.}
Reconstruction evaluates whether synthetic events train a model to invert events into images.
Optical flow evaluates a different property: whether event timing and motion structure support motion estimation.
The MVSEC experiment provides this complementary motion-estimation test.
MVSEC provides synchronized event data, grayscale frames, and optical-flow ground truth for standard event-vision evaluation \cite{mvsec2018,zhu2018evflownet}.

\paragraph{EV-FlowNet-style protocol.}
We use an EV-FlowNet-style recipe \cite{zhu2018evflownet} with fixed network architecture, event representation, losses, optimizer settings, augmentation, and 300k-step schedule.
For each simulated source, one model is trained on \texttt{outdoor\_day2} and evaluated on the official \texttt{outdoor\_day1} intervals and \texttt{indoor\_flying1}, \texttt{indoor\_flying2}, and \texttt{indoor\_flying3} at $dt=1$ and $dt=4$.
This common training split leaves the event source as the controlled variable across outdoor and indoor tests.
Each optical-flow setting is trained once; the tables report transfer performance from that run.

\paragraph{Implementation alignment.}
Our implementation follows the released EV-FlowNet design in the components that affect the simulator comparison: four encoder stages, two residual blocks, decoder skip connections, multi-scale flow heads, Charbonnier photometric loss, image warping, and smoothness regularization.
The input representation uses four event-image channels: positive counts, negative counts, latest positive timestamp, and latest negative timestamp.
Training samples use random temporal gaps from the same gap set for every event source, and evaluation uses fixed frame skips.
Grayscale APS images are kept in the official $0$--$255$ range used by the photometric loss.

\paragraph{Real-event reference.}
The model trained on real MVSEC events provides a reference scale under the same evaluation intervals.
The full numerical comparison is reported in the main paper; this supplement gives the protocol and additional qualitative examples.

\paragraph{Flow metrics.}
Following the EV-FlowNet-style reporting protocol \cite{mvsec2018,zhu2018evflownet}, average endpoint error is computed over event-active pixels with valid optical-flow labels as:
\begin{equation}
  \mathrm{AEE} =
  \frac{1}{|\Omega_v|}
  \sum_{(x,y)\in\Omega_v}
  \left\|
    \hat{\mathbf{u}}(x,y)-\mathbf{u}(x,y)
  \right\|_2,
  \label{eq:supp_aee}
\end{equation}
where $\Omega_v$ is the intersection of pixels with valid nonzero MVSEC flow labels and pixels containing at least one event in the evaluated interval.
The outlier percentage is:
\begin{equation}
  \mathrm{Out.} =
  \frac{100}{|\Omega_v|}
  \sum_{(x,y)\in\Omega_v}
  \mathbb{1}
  \left[
    e(x,y)>3\;\mathrm{px}
    \land
    e(x,y)>0.05\,\|\mathbf{u}(x,y)\|_2
  \right],
  \label{eq:supp_flow_outlier}
\end{equation}
where $e(x,y)=\|\hat{\mathbf{u}}(x,y)-\mathbf{u}(x,y)\|_2$.
AEE measures average error magnitude; the outlier percentage measures the prevalence of large errors.
\Cref{tab:supp_mvsec_protocol} lists the shared training and evaluation settings.

\begin{table}[!htbp]
  \centering
  \caption{\textbf{MVSEC optical-flow protocol.}
  Training and evaluation settings shared across event sources.}
  \label{tab:supp_mvsec_protocol}
  \scriptsize
  \setlength{\tabcolsep}{3pt}
  \begin{adjustbox}{max width=\linewidth}
  \begin{tabular}{lM{0.58\columnwidth}}
    \toprule
    Component & Fixed setting \\
    \midrule
    Training split & \texttt{outdoor\_day2}, left camera, one model per event source; the same trained model is used for outdoor and indoor evaluation. \\
    Training length & 300k optimizer steps for every source. \\
    Batch and crop & Batch size 8 with $256\times256$ random crops. \\
    Optimizer & Adam, learning rate $10^{-5}$, weight decay $10^{-4}$. \\
    LR schedule & Multiplicative decay 0.8 every four epochs. \\
    Event image & Positive count, negative count, latest positive time, latest negative time. \\
    Image scale & APS grayscale images kept in $0$--$255$ range for photometric loss. \\
    Frame gaps & Shared random training gaps from $\{2,4,6,8,10,12\}$; fixed $dt=1$ and $dt=4$ evaluation. \\
    Augmentation & Random temporal gap, horizontal flip, and random crop; no random rotation. \\
    Evaluation & \texttt{outdoor\_day1} official interval plus \texttt{indoor\_flying1}, \texttt{indoor\_flying2}, and \texttt{indoor\_flying3}, using the same \texttt{outdoor\_day2}-trained model for all targets. \\
    Metrics & Average endpoint error and outlier percentage. \\
    \bottomrule
  \end{tabular}
  \end{adjustbox}
\end{table}

\paragraph{DAVIS346-series sensor setting.}
The \method\ optical-flow experiment uses the DAVIS346-series setting calibrated from 20 activity-stratified MVSEC \texttt{outdoor\_day2} anchor windows and eight context windows (two each from \texttt{outdoor\_day1} and \texttt{indoor\_flying1/2/3}), with $\alpha=0.72$, $M=6$, $\tau_{\mathrm{ref}}=0.0025$ s, $\thetaon=0.45$, and $\thetaoff=0.50$.
The calibration objective matches real-event density while using local count, temporal histogram, polarity, and time-surface statistics.
No flow labels or EV-FlowNet feedback are used for parameter selection.
This setting represents the DAVIS346-series sensor regime used by the optical-flow comparison.

\paragraph{Sequence-level MVSEC reporting.}
The main paper reports public and calibrated AEE and outlier percentages for each MVSEC sequence at $dt=1$ and $dt=4$ under the common \texttt{outdoor\_day2} training split.

\begin{figure}[p]
  \centering
  \includegraphics[width=0.80\textwidth]{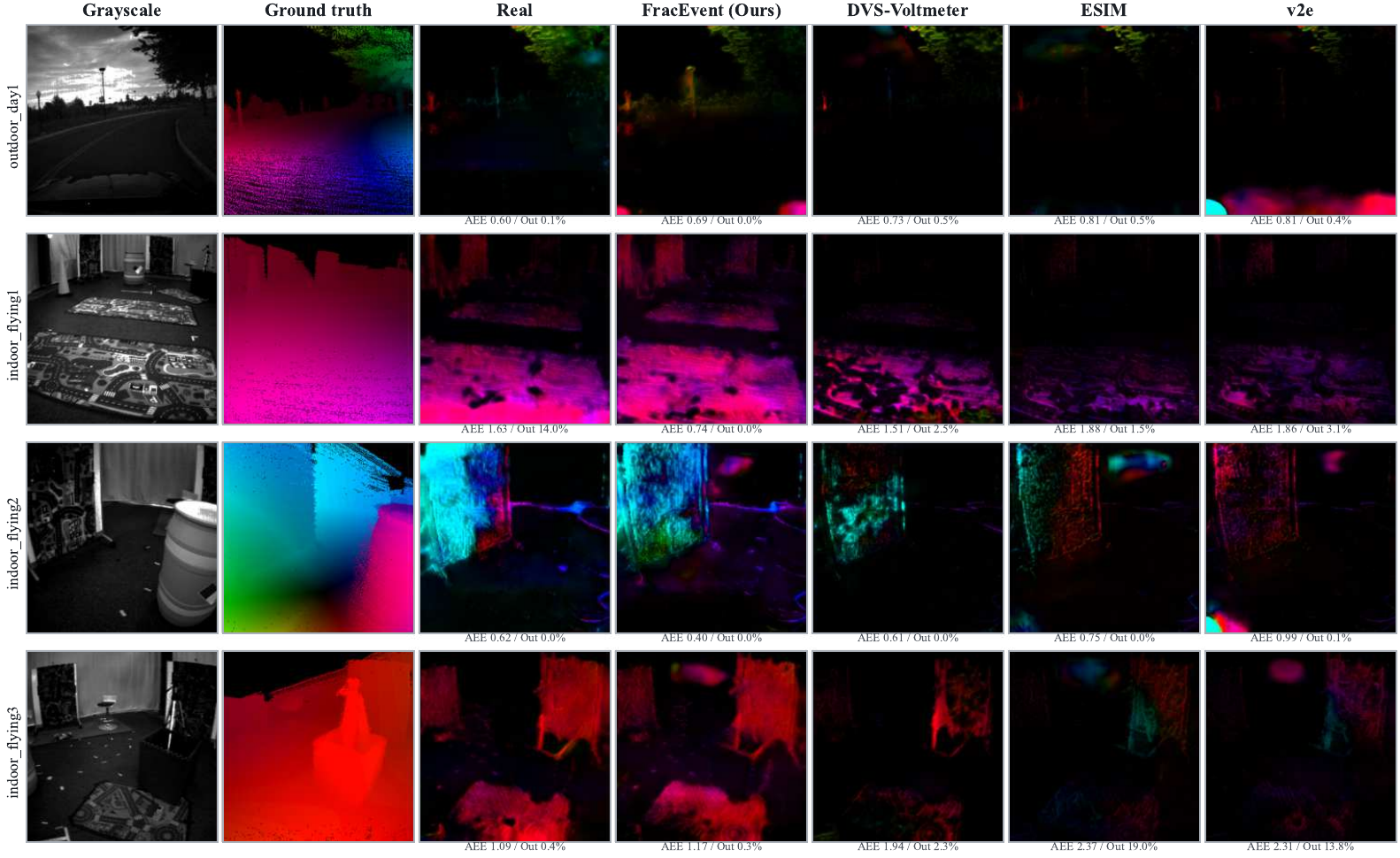}
  \caption{\textbf{Additional MVSEC optical-flow comparisons at $dt=1$.}
  Rows show the evaluated MVSEC sequences; columns compare the real event control and synthetic training sources under the same optical-flow transfer protocol.}
  \label{fig:supp_flow_dt1}
\end{figure}

\begin{figure}[p]
  \centering
  \includegraphics[width=0.80\textwidth]{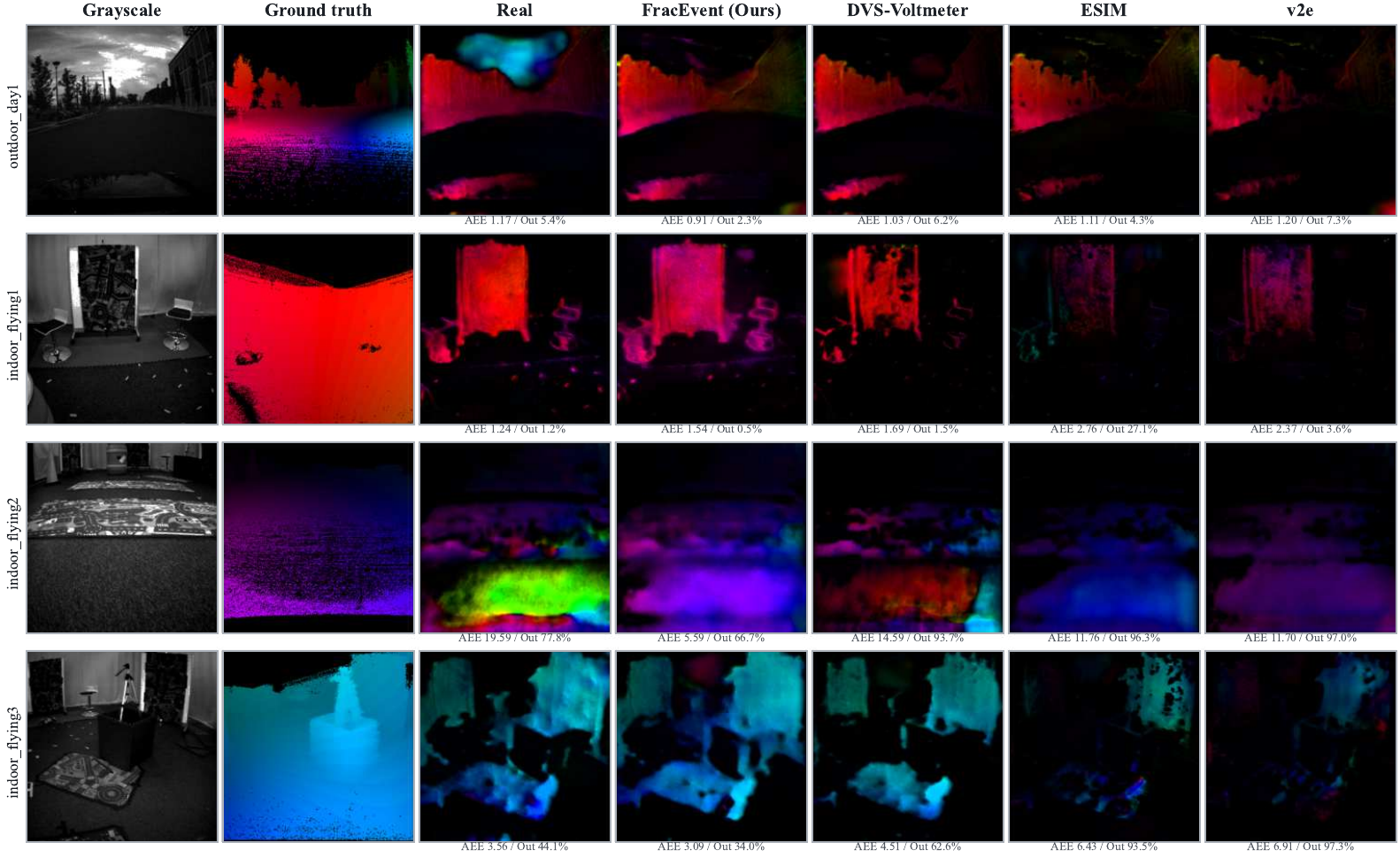}
  \caption{\textbf{Additional MVSEC optical-flow comparisons at $dt=4$.}
  Longer-interval examples corresponding to the $dt=4$ evaluation in the main table.}
  \label{fig:supp_flow_dt4}
\end{figure}

\paragraph{Additional optical-flow examples.}
\Cref{fig:supp_flow_dt1,fig:supp_flow_dt4} show additional public-setting MVSEC comparisons under the fixed EV-FlowNet-style protocol.
Both supplementary figures use the same target sequences, metric definitions, and flow-color convention as the main optical-flow comparison.
The displayed intervals are \texttt{outdoor\_day1} 10342--10346 and 10708--10712, \texttt{indoor\_flying1} 1103--1107, and \texttt{indoor\_flying2} 1333--1337; the corresponding 95th-percentile flow magnitudes are 4.36, 3.51, 6.23, and 16.33.

\FloatBarrier

\section{Computational Cost}
\label{app:cost}

\paragraph{Protocol.}
We process all APS frames in each of the four held-out DAVIS240C sequences at $240\times180$ resolution, comparing the selected four-mode FracEvent configuration ($\alpha=0.76$) with the fixed first-order control ($M=1$, $\alpha=1$).
Both use $\tau_{\mathrm{ref}}=0.0025$\,s, ON/OFF thresholds $0.5625/0.4$, 12 bisection steps, float32 states, and no parameter search.
The hardware is the RTX 5070 Ti workstation described in Sec.~B, using Python 3.11.15, PyTorch 2.11.0+cu130, CUDA 13.0, and one PyTorch CPU thread.
Each model receives an untimed one-interval warm-up, after which its state is discarded and a new instance starts from the sequence's first frame.
CUDA synchronization brackets the timed region.
Each reported measurement covers one complete timed pass per model and sequence.

The wall time includes initialization, frame transfer to the GPU, continuous-time event generation, event transfer to the CPU, and sorting within each APS interval.
Input decoding, evaluation metrics, disk output, and concatenation/sorting of the entire sequence into one array are excluded.
Event chunks are counted and discarded after each interval, as in a streaming consumer.
Generated event counts match the corresponding full-sequence ablation results for both models in every sequence.

\begin{table}[!htbp]
\centering
\caption{\textbf{Full-sequence streaming cost.} One measurement per row on a shared Windows GPU. Frames include the initial frame; throughput counts the remaining frame intervals. GPU memory denotes peak live tensor allocation reported by PyTorch.}
\label{tab:supp_cost}
\small
\setlength{\tabcolsep}{4pt}
\begin{tabular}{@{}llrrrrr@{}}
\toprule
Sequence & Model & Frames & Events (M) & Time (s) & Intervals/s & GPU (MiB) \\
\midrule
boxes\_rotation & FracEvent (4) & 1299 & 54.46 & 256.6 & 5.06 & 6.30 \\
boxes\_rotation & First-order (1) & 1299 & 61.99 & 272.7 & 4.76 & 3.36 \\
dynamic\_translation & FracEvent (4) & 1271 & 33.91 & 217.8 & 5.83 & 6.36 \\
dynamic\_translation & First-order (1) & 1271 & 37.18 & 183.3 & 6.93 & 3.18 \\
poster\_rotation & FracEvent (4) & 1357 & 67.70 & 263.4 & 5.15 & 6.32 \\
poster\_rotation & First-order (1) & 1357 & 76.25 & 281.6 & 4.81 & 3.38 \\
slider\_far & FracEvent (4) & 137 & 6.31 & 30.5 & 4.46 & 5.25 \\
slider\_far & First-order (1) & 137 & 7.04 & 33.0 & 4.12 & 2.82 \\
\bottomrule
\end{tabular}
\end{table}

\paragraph{Interpretation.}
The measured wall times characterize the current PyTorch implementation on a shared Windows workstation with concurrent GPU activity.
Runtime includes modal state updates, threshold localization, event transfer, and sorting.
The two configurations process identical input frames with the same thresholds, but produce different event workloads, as shown by the event counts in Table~\ref{tab:supp_cost}.
Their runtime differences therefore reflect both the temporal dynamics and the resulting event-processing workload.
The persistent mode array contains $MHW$ float32 entries (0.659\,MiB for four modes versus 0.165\,MiB for one at this resolution); temporary tensors, the reference array, and other allocations account for the larger measured GPU peaks.
The reported resource measurements cover wall time and peak live GPU tensor allocation; host RAM and end-to-end disk throughput are outside this measurement scope.
Table~\ref{tab:supp_cost} provides the computational costs corresponding to the fixed-parameter accuracy comparison in Sec.~D.
\section{Further Evaluation}
\label{app:scope}

\paragraph{Signal-level evaluation.}
The event-stream evaluation uses Count, IEI, Polarity, Time Surface, Spatial, Voxel, and Repeated IEI to characterize selected temporal, spatial, and polarity properties.
PECS evaluates simulated and real streams at the signal level \cite{pecs2024}, while the asynchronous spatiotemporal spike metric jointly incorporates spatial, temporal, and polarity structure through a kernel representation \cite{li2023spikemetric}.
Their protocols would specify spatial/temporal registration, measurement windows, and kernel scales independently of test outcomes.

\paragraph{Further task validation.}
The current downstream evaluation covers reconstruction and optical flow.
Neuromorphic object detection \cite{li2026neuromorphicdetection} and event-language understanding such as EventGPT \cite{liu2025eventgpt} are useful future tests of whether the generated streams preserve semantics beyond these low-level tasks.
\FloatBarrier

\clearpage
\twocolumn
\clearpage
{
    \small
    \renewcommand{\refname}{Bibliography}
    \bibliographystyle{ieeenat_fullname}
    \bibliography{main}
}

\end{document}